\documentclass[lettersize,journal]{IEEEtran}

\makeatletter
\long\def\@makecaption#1#2{\ifx\@captype\@IEEEtablestring%
\footnotesize\begin{center}{\normalfont\footnotesize #1}\\
{\normalfont\footnotesize\scshape #2}\end{center}%
\@IEEEtablecaptionsepspace
\else
\@IEEEfigurecaptionsepspace
\setbox\@tempboxa\hbox{\normalfont\footnotesize {#1.}~~ #2}%
\ifdim \wd\@tempboxa >\hsize%
\setbox\@tempboxa\hbox{\normalfont\footnotesize {#1.}~~ }%
\parbox[t]{\hsize}{\normalfont\footnotesize \noindent\unhbox\@tempboxa#2}%
\else
\hbox to\hsize{\normalfont\footnotesize\hfil\box\@tempboxa\hfil}\fi\fi}
\makeatother

\usepackage{amsmath,amsfonts}
\usepackage{algorithm}
\usepackage[noend]{algpseudocode}

\usepackage{array}
\usepackage[caption=false,font=normalsize,labelfont=scriptsize,textfont=scriptsize]{subfig}
\usepackage{textcomp}
\usepackage{stfloats}

\usepackage{url}
\usepackage{graphicx}
\usepackage{cite}
\usepackage{xcolor}

\begin{document}

\title{Privacy-preserving Universal Adversarial Defense\\ for Black-box Models}

\author{Qiao Li, Cong Wu, Jing Chen, Zijun Zhang, Kun He, Ruiying Du, Xinxin Wang, Qingchuang Zhao, Yang Liu
\thanks{Qiao Li, Jing Chen, Zijun Zhang, Kun He, Ruiying Du, and Xinxin Wang are with the  Key Laboratory of Aerospace Information Security and Trusted Computing, Ministry of Education, School of Cyber Science and Engineering, Wuhan University, Wuhan, China.
Email: \{liqiaoqiao233,chenjing,zijunzhang,hekun,duraying,xinelwang\}@whu.edu.cn}
\thanks{Qingchuan Zhao is with  the Department of Computer Science at the City University of Hong Kong (CityU), Hong Kong SAR. Email: cs.qczhao@cityu.edu.hk}
\thanks{Cong Wu and Yang Liu are with Cyber Security Lab, College of Computing and Data Science, Nanyang Technological University, Singapore. Email: \{cong.wu,yangliu\}@ntu.edu.sg}
}


\IEEEpubid{}

\thispagestyle{empty}
\maketitle

\begin{abstract}
Deep neural networks (DNNs) are increasingly used in critical applications such as identity authentication and autonomous driving, where robustness against adversarial attacks is crucial. These attacks can exploit minor perturbations to cause significant prediction errors, making it essential to enhance the resilience of DNNs. Traditional defense methods often rely on access to detailed model information, which raises privacy concerns, as model owners may be reluctant to share such data. In contrast, existing black-box defense methods fail to offer a universal defense against various types of adversarial attacks. To address these challenges, we introduce DUCD, a universal black-box defense method that does not require access to the target model's parameters or architecture. Our approach involves distilling the target model by querying it with data, creating a white-box surrogate while preserving data privacy. We further enhance this surrogate model using a certified defense based on randomized smoothing and optimized noise selection, enabling robust defense against a broad range of adversarial attacks. Comparative evaluations between the certified defenses of the surrogate and target models demonstrate the effectiveness of our approach. Experiments on multiple image classification datasets show that DUCD not only outperforms existing black-box defenses but also matches the accuracy of white-box defenses, all while enhancing data privacy and reducing the success rate of membership inference attacks.
\end{abstract}


\section{Introduction}
\label{sec:intro}

\IEEEPARstart{D}{eep} neural networks (DNNs) are widely applied across various fields, including large language models~\cite{lin2023pushing,lin2024splitlora}, and federated learning~\cite{lin2024adaptsfl,lin2023fedsn,fang2024automated}.
However DNNs have demonstrated vulnerability to adversarial examples, which are inputs with subtle perturbations designed to cause incorrect predictions~\cite{szegedy2013intriguing}. 
This vulnerability poses significant risks in safety-critical applications such as 
 biological identification~\cite{wu2020caiauth,wu2024s,wu2020liveness,wu2022echohand} and network traffic ~\cite{lin2024efficient,lin2024split,wu2024wafbooster,liang2024towards,liang2024ponziguard}, where adversarial examples can lead to severe consequences, including identity theft and traffic accidents. 
To mitigate these risks, various defense methods have been proposed, broadly categorized into empirical and certified defenses~\cite{zhangrobustify}. 
Empirical defenses, such as adversarial detection~\cite{aldahdooh2022adversarial} and adversarial training~\cite{qian2022survey}, have been effective in enhancing DNN robustness; 
however, they remain susceptible to adaptive attacks and lack provable robustness guarantees. 
In contrast, certified defenses offer a provable guarantee of robustness by ensuring that, 
within a certified $\ell_p$ radius (e.g., $\ell_1$, $\ell_2$, or $\ell_\infty$), 
adversarial attacks cannot successfully perturb the model’s predictions~\cite{cohen2019certified}.



\textbf{Existing method and research gap.} 
Although certified defense methods have demonstrated notable effectiveness~\cite{raghunathan2018certified}, the majority of them require access to white-box models. However, such prior knowledge is not always available to defenders, particularly third-party entities, due to privacy and security concerns from model owners. Implementing these defenses typically demands substantial resources, including high-performance GPUs and expert knowledge, making them less feasible in practice. Consequently, recent studies have focused on providing provable robustness in black-box settings, where defenders must offer a robustness-certified model based on limited information from model owners. Existing methods, such as those proposed by Teng et al.~\cite{teng2020ell_1} and Salman et al.~\cite{salman2020denoised}, employ Gaussian noise for denoised smoothing based on surrogate models. However, these approaches fall short in practical black-box defense scenarios, as they assume that the target model adheres to a specific structure. To address this gap, Zhang et al.~\cite{zhangrobustify} utilize query-based techniques and zero-order optimization to estimate information for denoised smoothing. Unfortunately, these methods only achieve certified defenses under $\ell_2$-norm constraints, making them vulnerable to $\ell_\infty$-norm attacks, which are prevalent in real-world scenarios. Moreover, they offer a restricted certified radius, further limiting their practical effectiveness. This highlights the need for more general and practical black-box certified defenses that can overcome these limitations.
%

\textbf{DUCD.} 
In this paper, we present a practical and effective universal defense method designed to address two critical properties: (i) \emph{Model-agnosticism}, which ensures that the method can be applied to any target model without requiring prior knowledge, and (ii) \emph{Norm-universality}, which provides provable robustness against attacks across different $\ell_p$-norms. A key requirement for such a defense method is to achieve the highest possible certified radius to enhance its practical effectiveness. However, existing black-box defenses have struggled to simultaneously meet the criteria of model-agnosticism and norm-universality. To bridge this gap, we propose a novel universal defense approach that maximizes the certified radius while maintaining model privacy in black-box settings. Our method requires only query access to the target model to construct a certified classifier. The core innovation of our approach lies in applying randomized smoothing~\cite{cohen2019certified}, a technique that typically provides norm-universality and a high certified radius, without the need for access to the target model's internal details. This distinguishes our method from denoised smoothing techniques~\cite{salman2020denoised}, which are less flexible and offer lower certified radii. Recognizing that traditional randomized smoothing methods often rely on prior knowledge of the target model, we overcome this limitation by designing a target loss function that aligns the surrogate model's outputs with those of the target model, ensuring functional consistency while preserving the target model's privacy. To further enhance the effectiveness of our defense, we optimize the certified radius by treating input noise as a tunable variable, resulting in a significantly higher certified radius compared to conventional randomized smoothing applied directly to surrogate models.

Our method offers three distinct advantages over previous certified defenses.
(i) \emph{Universal defense}: Our approach is both model-agnostic and norm-universal, providing provable robustness against $\ell_p$-norm attacks without requiring prior knowledge of the target model~\cite{salman2020denoised,teng2020ell_1}.
(ii) \emph{Privacy-preserving}: Our method enhances privacy-preservation by generating a surrogate model that significantly reduces the attack success rate. For both the surrogate model and its certified classifiers, the attack success rate is lowered to that of a random guess~\cite{hammoudeh2024provable,raghunathan2018certified}.
(iii) \emph{High certified radius}: By fine-tuning noise parameters, our method achieves a competitive certified radius, surpassing the maximum certified radius obtained through randomized smoothing with Gaussian noise~\cite{cohen2019certified}.
Our main contributions are summarized as follows:
\begin{itemize}	
    \item To the best of our knowledge,
    we are the first to achieve a practical defense for black-box models, which is model-agnostic, norm-universality, privacy-preserving, and characterized by a high certified radius.
    
    \item We introduce a novel universal defense method for black-box models. Our approach involves generating a surrogate model using a query-based technique, designing a targeted loss function to ensure functional consistency, and optimizing certified defense through randomized smoothing. This method establishes robust certified classifiers with a high certified radius, effectively protecting against adversarial attacks across any $\ell_p$-norm.

    \item We demonstrate the effectiveness and robustness of our approach through extensive experiments and comparisons with state-of-the-art defense techniques. Our results indicate a significant improvement in certified accuracy over existing black-box defenses by more than 20\%, while also delivering competitive performance compared to white-box defenses.
\end{itemize}

\section{Preliminaries}
\subsection{Randomized Smoothing}
Randomized smoothing introduces random noise or perturbations to input samples, ensuring that minor changes in inputs do not alter the predictions of the DNNs. By randomly sampling input space and averaging the results, smoothed predictions are obtained. This averaging minimizes the impact of noise and perturbations on DNN predictions, enhancing their robustness.
The essence of randomized smoothing is to construct a smoothed classifier from the target classifier. 
Given a target classifier $f(x):\mathbb{R}^d \rightarrow\mathcal{Y}$, where $\mathbb{R}^d$ represents the input space and $\mathcal{Y}$ indicates the output space, a randomly smoothed classifier, $g$, is defined as:
\begin{equation}
	g(x) =\mathop{\arg\max}\limits_{c\in\mathcal{Y}}\ \mathbb{P}(f(x+\epsilon)), \quad \epsilon\sim\mathcal{N}(0,\sigma^2 I).
\end{equation}

For a given input $x$, the smoothed classifier $g$ outputs the class probabilities that the target classifier $f$ is most likely to predict in the presence of random noise $\epsilon\sim\mathcal{N}(0,\sigma^2 I)$.
The target classifier $f$ outputs the most likely category $c_A$ for input x with probability $p_A$, and outputs the next most likely category with probability $p_B$.
In the case of a classification task, for $\epsilon \sim \mathcal{N}(0,\sigma^2 I)$, assuming $c_A \in \mathcal{Y}$ and $p_A$, $p_B \in \left[0, 1\right]$, we have:

\begin{equation}
\mathbb{P}(f(x+\epsilon) = c_A) \geq p_A \geq p_B \geq \mathop{\max}\limits_{c \neq c_A}\ \mathbb{P}(f(x+\epsilon) = c).
\end{equation}

Replacing $p_A$ with a lower bound $\underline{p_A}$ and $p_B$ with an upper bound $\overline{p_B}$ in the above equations still holds.

\subsection{Robustness Guarantee} 
Let $p_A$ denote the probability that the target classifier $f$ outputs the most probable class $c_A$, and $p_B$ denote that of the second most probable class $c_B$. The certified radius is the minimum $\ell_p$ norm of a perturbations, $\delta$, that satisfies the following \emph{robustness boundary conditions}~\cite{cohen2019certified}:

\begin{equation}
\label{eq:radius_1}
\begin{gathered}
\mathbb{P}\left(\frac{\mu_x(x-\delta)}{\mu_x(x)} \leq t_A\right)=\underline{p_A}, \\
\mathbb{P}\left(\frac{\mu_x(x-\delta)}{\mu_x(x)} \geq t_B\right)=\overline{p_B},\\
\mathbb{P}\left(\frac{\mu_x(x)}{\mu_x(x+\delta)} \leq t_A\right)=\mathbb{P}\left(\frac{\mu_x(x)}{\mu_x(x+\delta)} \geq t_B\right),\\
t_A = \Phi^{-1}(\underline{p_A}), t_B = \Phi^{-1}(\overline{p_B}).
\end{gathered}
\end{equation}

Here, $t_A$ and $t_B$ are auxiliary parameters that satisfy Eq. (2). $\Phi^{-1}$ is the inverse function of $\mu_x(x-\delta)$/$\mu_x(x)$. The certified radius of the smoothed classifier $g$ in the $\ell_p$ norm is:

\begin{equation}
R = \frac{\sigma}{2} \cdot \left(t_A -t_B \right).
\end{equation}

Therefore, for any $\lvert\lvert \delta \rvert\rvert _p < R$, we have $g(x+\delta)=c_A$. This means that for any possible target classifier $f$ and any possible $x$, the output of the smoothed classifier $g$ will not exceed $P(c_A = g(x)) \pm R$.

As a result, when the noise level $\sigma$ is high or the probability value of the maximum class $c_A$ is large, or the probability values of other classes are low, the robustness radius $R$ will increase. When the robustness radius $R\rightarrow \infty$, we have $p_A\rightarrow1$ and $p_B\rightarrow 0$. Since the Gaussian distribution exists in the entire input space $\mathbb{R}^d$, there are cases with $p_A=1$ where $f(x+\epsilon)=c_A$.

\subsection{Threat Model}
We outline the threat model for adversarial attacks, covering the attacker's goals and knowledge of the target model. 
We consider a malicious third party as an attacker. The attacker acquires relevant prior knowledge of the model and crafts an adversarial example using this knowledge. The attacker selects a sample for manipulation, initiating subtle modifications, aiming to misclassify it. We next delineate the attacker's goals and knowledge.

The attacker's main goal is to generate adversarial examples using prior knowledge of the model, guiding the target model to make incorrect predictions and thus degrading its performance. 
More specifically, the attacker’s goal can be defined as the perturbation goal and the target effectiveness goal  as follows:
\begin{itemize}
    \item \emph{Goal - \expandafter{\romannumeral 1}: perturbation goal.} This goal implies that the perturbations added to the sample should be hard to detect. It is intended to minimize the added perturbations.
    \item \emph{Goal - \expandafter{\romannumeral 2}: target effectiveness goal.} 
This goal pertains to the model's misclassification behavior concerning the adversarial example during testing, aiming to ensure successful misclassification of the target model.
\end{itemize}


We consider a white-box attack where the attacker possesses specific knowledge. Firstly, the attacker can choose the samples to target. Secondly, the attacker has prior knowledge of the target model, including its structure, parameters, and training data. Additionally, the attacker is capable of adaptive attacks. This means that the attacker, upon discovering the implementation of a defense by the target model, can modify their strategy and launch a new attack against the updated defense. It's important to note that this adaptive attack remains a white-box attack.

\section{Methodology}
In this section, we present overview of methodogy and detail each step.
\subsection{Overview}
In this section, we present a Distillation-based Universal Black-box model Defense (DUCD). The key idea is to employ queries to generate a surrogate model of the target model, thereby converting the black-box scenario into a white-box one. Leveraging this surrogate model, we propose a certified defense strategy aimed at developing a universal defense.

Let $f(x)$ be a pre-trained black-box prediction model that maps an input $x$ to a predicted output. 
The black-box scenario studied in this paper is one where the owner of the model is unwilling to share model parameters and training data. 
Therefore, the only interaction with the black box is to submit an input and receive the corresponding prediction. 
Formally, given a black-box base model $f$, the goal is to devise a certified classifier $D$ exclusively using input-output queries $Q(f)$, where the $Q(f)$ operation provides the output label or logits of input $x$ for $f$, and derive a robust certified model $D(x)$ resilient to adversarial perturbations.
\subsection{Distillation-based Surrogate Model Generation }
The goal of the surrogate model $A$ is to accurately fit the predictions of the target black-box model $T$ on its input space $D_b$. Specifically, we aim to find the optimal parameter $\theta_a$ for the surrogate model $A$ such that the error between $A(x)$ and $T(x)$ is minimized for all $x \in D_b$. Then for all $x \in D_b$:\\
\begin{equation}
    \mathop{\arg\min}\limits_{\theta _a} \ \mathbb{P}_{x \sim D_b}
    \left[\arg\max\ T(x) \neq \arg\max \ A(x)\right].
\end{equation}

\begin{figure}[!t]
\centering
\setlength{\abovecaptionskip}{5pt} 
\includegraphics[width=\linewidth]{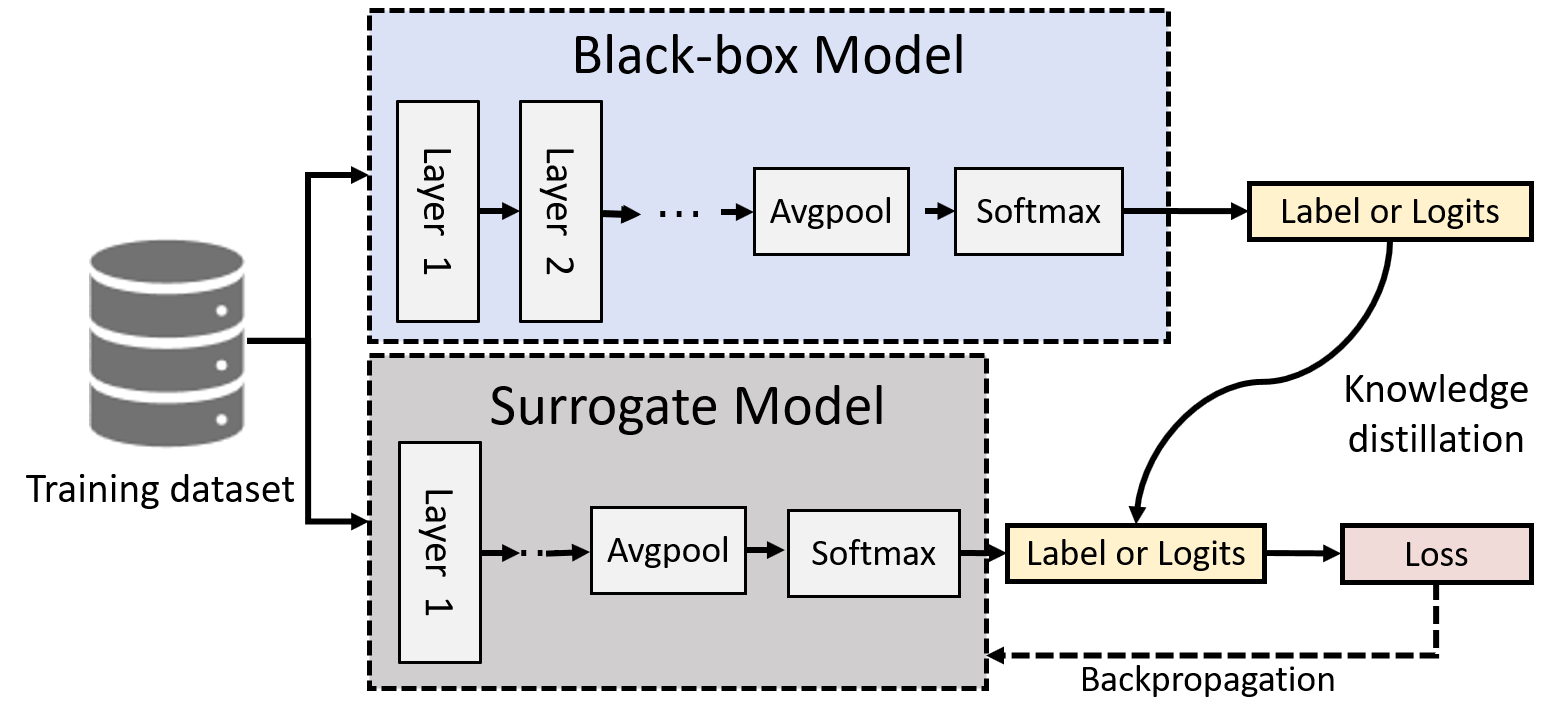} 
\captionsetup{justification=centering}
\caption{Illustration of knowledge distillation in DUCD.}
\label{fig:fig_DUCD}
\end{figure}

In order to achieve distillation-based surrogate model generation, DUCD leverages the idea of knowledge distillation, transferring knowledge from the target model to a white-box surrogate model \cite{hinton2015distilling}.
In DUCD, the target model serves as a teacher, from which the surrogate model learns as a student.
As illustrated in Fig. \ref{fig:fig_DUCD}, the logits generated by the teacher network are utilized as the training targets for the student network's outputs on the training dataset.

To address the issue of gradient vanishing during training, we utilize the $\ell_1$ norm loss as our loss function.
Despite the non-differentiability of the $\ell_1$ norm loss, empirical studies have demonstrated its resilience to gradient vanishing and its ability to yield improved results \cite{fang2019data}. 
The loss function is defined as follows:\\
\begin{equation}
Loss = \sum_{i=1}^{C} |t_i - a_i|, i\in {1,...,C}.
\end{equation}
where $t_i$ and $a_i$ are the logits belonging to the black box model and the surrogate model, and C is the number of training dataset categories. 

The surrogate model is iteratively trained on the dataset, using the predicted output of the black-box model as the target. This iterative approach aims to align the predictions of the surrogate model with those of the black-box model. Training termination is controlled based on the query budget constraints. After training, the surrogate model effectively mimics the black-box model's predictions.
\begin{figure}[!t]
    \centering
    \subfloat[Robust space: $\tau$]{\includegraphics[width=0.3\linewidth]{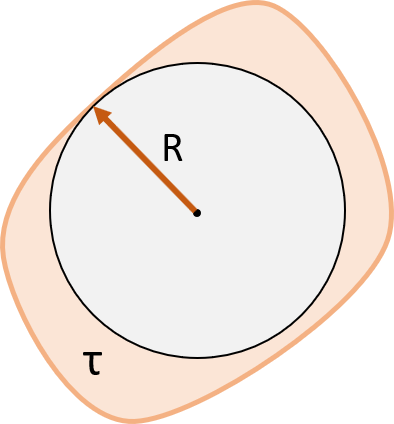} } \hspace{1mm}
    \subfloat[Scalar optimization]{\includegraphics[width=0.3\linewidth]{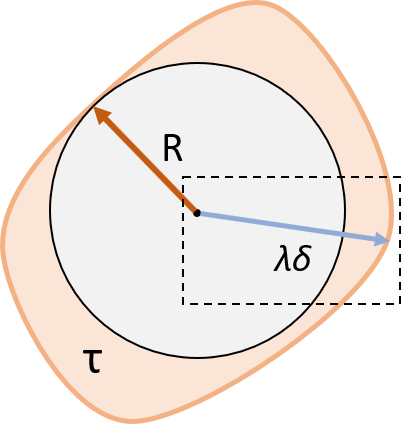} }\hspace{1mm}
    \subfloat[Direction optimization]{\includegraphics[width=0.3\linewidth]{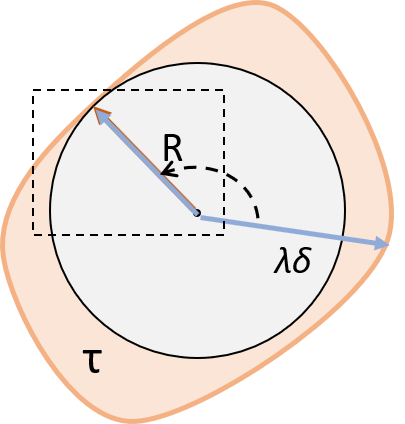} }
    \caption{Robust radius derivation.}
    \label{fig:robust-radius-derivation}
\end{figure}
\subsection{Universal Black-box Defense Using Surrogate Models}
\label{sec:wucd}
Universal defense necessitates both model-agnosticism and norm-universality. Additionally, it should yield a higher certified radius for any $\ell_p$ perturbation.
After obtaining the surrogate model through distillation, we apply randomized smoothing and noise selection to create a universal defense for the black-box model with a higher radius.
Randomized smoothing-based certified defense enhances classifier robustness against adversarial examples, reduces misclassification risk, and offers a level of robustness assurance.

\textbf{Robust radius.}
We utilize two-stage optimization \cite{hong2022unicr} to derive precise certified radii. As illustrated in Fig.\ref{fig:robust-radius-derivation}, the perturbation added to the input stays within the robust space $\tau$, maintaining the original prediction. The first stage ensures that $\delta$ aligns with the robustness bound, and the second stage aims to minimize the $\ell_p$ norm.
\begin{algorithm}[!h]
	\caption{Scalar optimization}
	\label{alg:scalar}
	\begin{algorithmic}[1]
		\Statex \textbf{Input} Lower bound of probability $\underline{p_A}$; upper bound of probability $\overline{p_B}$;  scalar value of perturbation $\lambda$;  perturbation $\delta$; probability density function of the noise  $\mu_x$; Monte Carlo samples' number $n$;  $K$'s threshold $K_m$; dichotomous search iterations number $N$.
		\Statex \textbf{Output} The scalar $\lambda$ that minimizes $|K|$.

		\State  Initial scalars $\lambda_a$ and $\lambda_b$ such that $K > 0$ and $K < 0$.
		\State $\lambda = (\lambda_a + \lambda_b)/2$.
		\While{$N > 0$ and $|K| > K_m$}
		\If {$K > 0$}
		\State $\lambda_a = \lambda$.
		\Else
		\State $\lambda_b = \lambda$
		\EndIf
		\State $\lambda = (\lambda_a + \lambda_b)/2$.
		\State Compute $K$ with $\lambda$ .
		\State $N = N - 1$.
		\EndWhile
		\State \textbf{return} $\lambda$.
	\end{algorithmic}
\end{algorithm}


The first stage utilizes scalar optimization to find a scaling factor $\lambda$ that adjusts the perturbation $\delta$ to the robust boundary. Let $|K|$ represent the distance between the perturbation $\delta$ and the robustness boundary. In this paper, we use a binary search method to find the scaling factor that minimizes $|K|$. When $K=0$, the perturbation $\delta$ lies exactly on the robustness boundary. By fixing the direction of $\delta$, it is necessary to find two scalars such that $K>0$ and $K<0$. First, we compute $K$ based on the scalar $\lambda_a$. If $K>0$, then the scaled perturbation $\lambda_a\delta$ lies within the robustness boundary. By adjusting the scalar, we can find $\lambda_b$ such that $K<0$, and vice versa. In the proposed approach, $K$ is iteratively computed with $\lambda = \frac{1}{2} (\lambda_a + \lambda_b)$. If $K>0$, then let $\lambda_a = \lambda$; otherwise, let $\lambda_b = \lambda$. This iterative process is repeated until $K$ is less than a threshold or the number of iterations reaches a specified limit. The steps are detailed in Algorithm \ref{alg:scalar}.

In the second stage, direction optimization is performed to optimize the direction of $\delta$ to minimize $||\lambda\delta||_p$ . We employ Particle Swarm Optimization (PSO) to determine the optimal $\theta$ that minimizes the $\ell_p$ norm within the robustness bound. Within each PSO iteration, the particle's position is denoted by $\delta$, with a cost function $f_{PSO}(\delta) = ||\lambda\delta||_p$. Here, $\lambda$ is derived from scalar optimization. PSO seeks the optimal $\delta$ to minimize this function.
Hence, the intractable optimization problem in Eq. \ref{eq:radius_1} can be reformulated as follows:
\begin{align}
\label{eq:rr}
\nonumber R&=\|\lambda \delta\|_p, \\
\nonumber&\text{s.t. } \delta \in \underset{\delta}{\arg \min }\|\lambda \nonumber\delta\|_p, \quad \lambda=\underset{\lambda}{\arg \min }|K| \text{,} \\
\nonumber\mathbb{P}&\left(\frac{\mu_x(x-\lambda \delta)}{\mu_x(x)} \leq t_A\right)=\underline{p_A}, \\
\nonumber\mathbb{P}&\left(\frac{\mu_x(x-\lambda \delta)}{\mu_x(x)} \geq t_B\right)=\overline{p_B}, \\
\nonumber K&=\mathbb{P}\left(\frac{\mu_x(x)}{\mu_x(x+\lambda \delta)} \leq t_A\right)-\mathbb{P}\left(\frac{\mu_x(x)}{\mu_x(x+\lambda \delta)} \geq t_B\right), \\
t_A &= \Phi^{-1}(\underline{p_A}), t_B = \Phi^{-1}(\overline{p_B}).
\end{align}

Eq.~\ref{eq:rr} aims to scale $\delta$ using $\lambda$, moving it towards the robust boundary, and reducing $|K|$ to 0. 
By employing $\lambda$, the scaled $\delta$ effectively approaches the boundary. 

Simultaneously, direction optimization optimizes the direction of the perturbation $\delta$ to find the certified radius $R = ||\lambda\delta||_p$. Monte Carlo methods\cite{cohen2019certified} are used to estimate the bounds of probabilities $\underline{p_A}$ and $\overline{p_B}$.
Using the estimated $\underline{p_A}$, $\overline{p_B}$, noise probability density function, and perturbation $\delta$, we employ Monte Carlo methods to determine the $CDF$ of the ratio $\frac{{\mu_x(x - \lambda\delta)}}{{\mu_x(x)}}$.
We then calculate the auxiliary parameters $t_A$ and $t_B$ as the Algorithm \ref{alg:computation}
.
\begin{algorithm}[!b]
	\caption{The computation of $t_A$ and $t_b$}
	\label{alg:computation}
	\begin{algorithmic}[1]
		\Statex \textbf{Input}: Lower bound of probability $\underline{p_A}$; upper bound of probability $\overline{p_B}$;  scalar value of perturbation $\lambda$;  perturbation $\delta$; probability density function of the noise  $\mu_x$; Monte Carlo samples' number $n$.
		\Statex \textbf{Output}:  Auxiliary parameters $t_A$ and $t_B$. 
  
		\State Sample $n$ noise values, $\epsilon \in \mathbb{R}^{n \times d}$ , from the discrete probability density function.
		\State Use these $n$ noise samples, $\mu_x$, $\lambda$ and $\delta$ to compute $\mu_x(x - \lambda\delta)$.
		\State Use the Monte Carlo method to estimate the cumulative distribution function $\Phi$ for $\mu_x(x - \lambda\delta)$
		\State \textbf{return} $t_A = \Phi^{-1}(\underline{p_A})$ ;\quad $t_B = \Phi^{-1}(\overline{p_B})$
	\end{algorithmic}
\end{algorithm}

\textbf{Noise selection.}
Existing randomized smoothing methods typically use Gaussian noise, during the training of the smoothed classifier \cite{cohen2019certified}. However, different noise PDFs can result in different certified radii. To maximize the certified radius, we consider the noise PDF as a variable to be optimized.

Let $\mu$ be the noise PDF, we formulate the optimal noise selection problem as finding an optimal $\mu$ for the classifier to optimize the overall performance of the certified classifier. 
To further tune $\mu$ for each classifier, we represent $\mu$ as $\mu(x, \alpha)$, where $\alpha$ represents the set of hyperparameters to be tuned, i.e., $\alpha = [\alpha_1, \alpha_2, ..., \alpha_m]$. 
The noise selection procedure employs a grid search to find the best parameters of the noise PDF.


\begin{algorithm}[htb]
	\caption{Certified classifier}
	\label{alg:training}
	\begin{algorithmic}[1]
		\Statex \textbf{Input}: Train set $X$; label set $C$; abstain threshold $\zeta$; base classifier $f$; Monte Carlo sampling number $n$; certification threshold $\iota$ 
		
        \Statex \textbf{Output}: Certified classifier $g$
		\Statex \textbf{TRAIN}

		\For{$x_i$, $c_i$ in $X$, $C$}
		\State Calculate the class distribution $f(x_i+\epsilon)$ based on the labels $c_i$ of $x_i$
		\State Sample $n$ noise samples to approximate the class distribution $f(x_i+\epsilon)$
		\State Count the number of occurrences of each class in the sample $counts$
		\For{each $c$}
		\If{$c \neq c_A$ and $counts[c]>counts[c_A] *\zeta$}
		\State Update $g$
		\EndIf
		\EndFor
		\EndFor
		\State return $g$

		\Statex \textbf{PREDICT}

		\State Sample $n$ noise samples to approximate a class distribution $f(x+\epsilon)$
		\State Sample a example $x_i$ from the class distribution $f(x_i+\epsilon)$
		\For{ each $c$}
		\If{$c\neq c_A$ and $counts[c]>counts[c_A] * \zeta$}
		\State return $abstain$
		\EndIf
		\EndFor
		\State return $c_A$

		\Statex \textbf{CERTIFICATION}

		\State Sample $n$ noise samples to approximate the class distribution $f(x_i+\epsilon)$
		\State Count the number of occurrences of each class in the sample $counts$
		\State Take the class with the most occurrences in counts $c_A$
		\For{each $c$}
		\If{$c\neq c_A$ and $counts[c] > counts[c_A] * \zeta$}
		\State return $abstain$
		\EndIf
		\EndFor
		\State Calculate the frequency of class $c_A$ $counts[c_A] / n$
		\If{$counts[c_A] / n > \iota$}
		\State return $c_A$
		\Else
		\State return $abstain$
		\EndIf
	\end{algorithmic}
\end{algorithm}

\textbf{Certified classifier based on randomized smoothing.}
The certified defense based on randomized smoothing utilizes noise selection during training to approximate the input's category distribution. The prediction result is determined by selecting the most frequent category. If the occurrences of other categories exceed a predefined threshold, the classifier abstains from making a prediction. 
A randomized smoothing-based classifier for certified defense employs noise sampling in training to estimate the class distribution in the samples, ultimately predicting the class with the highest occurrence.
If the occurrence count of other classes surpasses the predefined threshold, the abstain result is returned.
The base classifier generates n samples by applying n different noise perturbations to input $x$.
The prediction will be $c_A$ if it appears more frequently than any other class.
The certified classifier employs hypothesis testing to adjust the abstain threshold, ensuring that the probability of incorrectly returning a class is restricted to $\alpha$, as depicted in Algorithm \ref{alg:training}.
The certification process evaluates the certification's success by comparing the frequency of the most common class with the certification threshold. This process is also outlined in Algorithm \ref{alg:training}.\\

\section{Experiments}
In this section, we provides a comprehensive analysis of our method's performance.

\subsection{Experimental Setup}
\textbf{Datasets.}
We evaluate our proposed DUCD method on three benchmark datasets: MNIST~\cite{lecun1998gradient}, SVHN~\cite{netzer2011reading}, and CIFAR10~\cite{krizhevsky2009learning} 
Furthermore, we split the CIFAR10 training set into two halves (dubbed CIFAR-S): one half is used to train the target model, while the other half is used for distillation.
CIFAR-S is designed for scenarios where the model owner is unwilling to disclose the original dataset, but an independent and identically distributed dataset is available. In addition, for scenarios where only a similar dataset is available, we employ CIFAR10.1~\cite{recht2018cifar} for distillation, which is similar to CIFAR10 but with a slightly different distribution.

\textbf{Metrics.}
The evaluation of the surrogate model involves assessing its absolute accuracy and relative accuracy. The absolute accuracy refers to the test accuracy of the surrogate model, while the relative accuracy represents the ratio of the test accuracy of the surrogate model to that of the target model. Additionally, we employ the 'approximate certified test set accuracy', commonly used in prior studies\cite{cohen2019certified,zhang2022adversarial,hong2022unicr,chen2022input}, to compute the certification radius for $\ell_p$-norm perturbations on input samples.
Certified accuracy is then measured as the percentage of correctly classified data for which the certified radius exceeds a specified threshold $R$.
Following~\cite{zhang2020black}, we evaluate the overall certified robustness by plotting the correlation curve between the certified accuracy and certified radius for noise sampled with varying $\sigma$, and computing the area under the curve:
\begin{equation}
\label{eq:score}
score = 
\int_0^{+\infty} \max _\sigma\left(Acc_\sigma(R)\right) d R, \sigma \in \Sigma.
\end{equation}

\textbf{Environments.}
All experiments are conducted on Ubuntu 20.04 with an Intel(R) Xeon(R) Gold 6133 CPU @ 2.50GHz and an NVIDIA GeForce RTX 4090 GPU.
The experiments are implemented in Python 3.8 with PyTorch 1.12.0, CUDA 12.0, and the adversarial-robustness-toolbox~\cite{art2018}.

\begin{table*}[!t]
\caption{Accuracy of surrogate model generation under different distribution.\label{tab:table1}}
\centering
\renewcommand{\arraystretch}{1.4} 
\begin{tabular}{|c|ccc|ccc|ccc|}
\hline

Target model& \multicolumn{3}{c|}{\textbf{CIFAR10}}& \multicolumn{3}{c|}{\textbf{CIFAR10.1}}& \multicolumn{3}{c|}{\textbf{CIFAR-S}}\\
\cline{2-4}\cline{5-7}\cline{8-10}
architecture& Target & Surrogate & Ratio& Target & Surrogate & Ratio& Target & Surrogate & Ratio\\
\hline
Resnet-18    &93.01  &92.74  &99.70 &93.01  &88.57  &95.22 &90.29  &89.93  &99.60\\
\hline
Resnet-34    &93.15  &93.01  &99.84 &93.15  &89.07  &95.62&89.92  &89.67  &99.72\\
\hline
Resnet-50   &93.56  &92.87  &99.26 &93.56  &87.90  &93.95 &90.27  &89.38  &99.01\\
\hline
\end{tabular}
\end{table*}

\subsection{Surrogate Model Evaluation}
Recent studies in knowledge distillation demonstrate that a relatively small student model can often perform on par with a larger teacher model~\cite{fang2019data,han2024effectiveness}. Therefore, we employ ResNet-18 as the surrogate model architecture in the following experiments.

\begin{figure}[!t]
\centering
\includegraphics[width=0.95\linewidth]{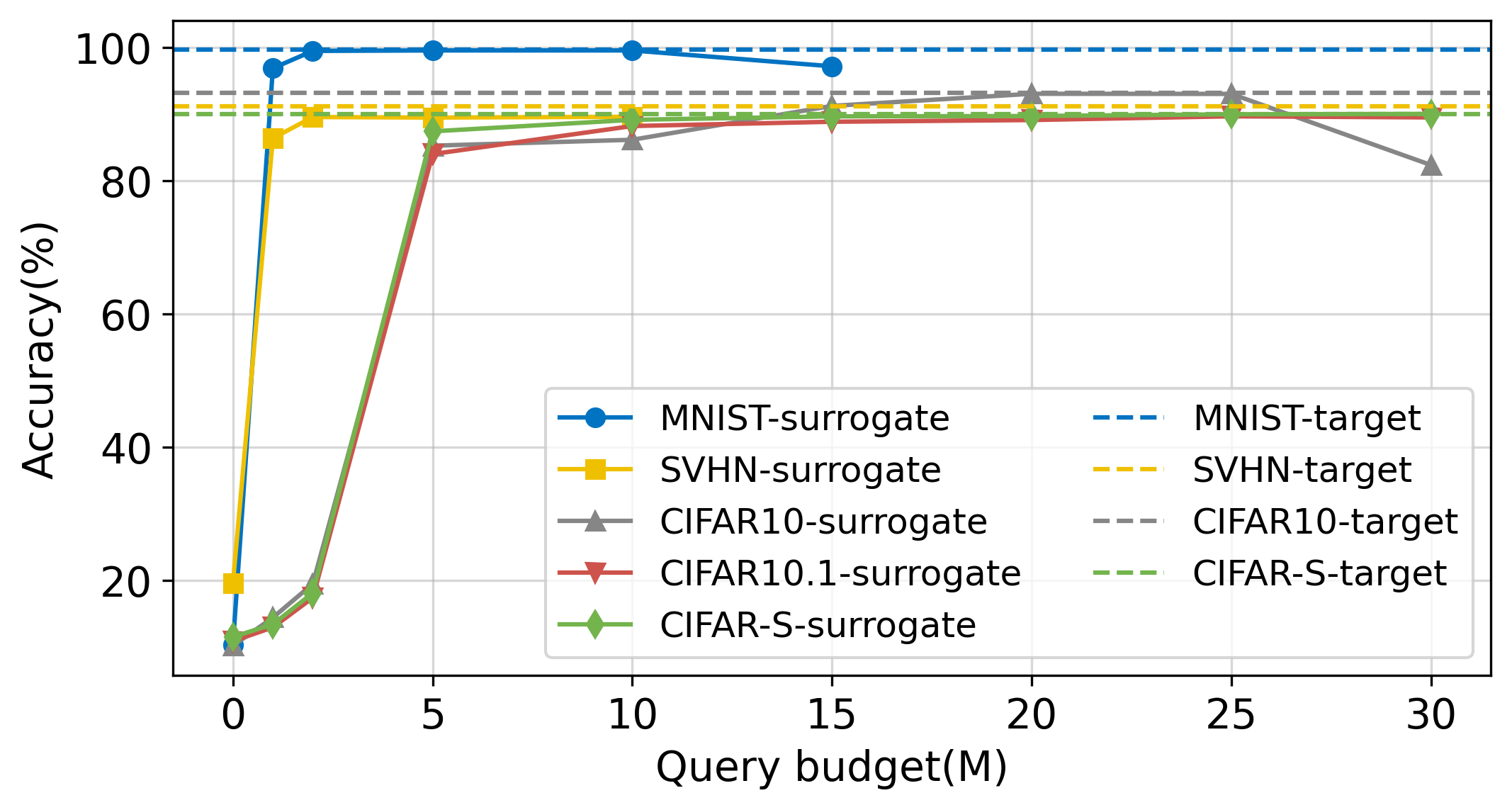} 
\caption{Accuracies of surrogate models under different query budgets.}
\label{fig:qurey-accuracy}
\end{figure} 
Firstly, we generate surrogate models for Resnet-34 classifiers under various query budgets. Fig. \ref{fig:qurey-accuracy} shows the accuracies of the resulting models, which serve as the basis for the subsequent experiments.
Hence, we use a default query budget of 2M for the MNIST and SVHN datasets, and 20M for the CIFAR10 dataset. 
Under these query budgets, we compare the effectiveness of the surrogate model generation schemes.
Tables \ref{tab:table1} and \ref{tab:table2} present the performance of different target model architectures on various datasets, as well as that of the surrogate models for comparison.
\begin{table}[thb]
\caption{Accuracy of surrogate model generation with different datasets.\label{tab:table2}}
\centering
\renewcommand{\arraystretch}{1.4} 

\resizebox{0.75\linewidth}{!}{
\begin{tabular}{|c|ccc|}
\hline
Target model & \multicolumn{3}{c|}{\textbf{MNIST}} \\
\cline{2-4}
architecture & Target & Surrogate & Ratio \\
\hline
Resnet-18    & 99.58  & 99.57  & 99.99 \\
\hline
Resnet-34    & 99.65  & 100.00 & 100.35 \\
\hline
Resnet-50    & 99.55  & 99.50  & 99.95 \\
\hline
\end{tabular}
}

\medskip

\resizebox{0.75\linewidth}{!}{
\begin{tabular}{|c|ccc|}
\hline
Target model & \multicolumn{3}{c|}{\textbf{SVHN}} \\
\cline{2-4}
architecture & Target & Surrogate & Ratio \\
\hline
Resnet-18    & 91.60  & 91.99  & 100.43 \\
\hline
Resnet-34    & 91.08  & 89.52  & 98.28 \\
\hline
Resnet-50    & 90.16  & 90.09  & 99.92 \\
\hline
\end{tabular}
}

\end{table}

The results show a slight decrease in accuracy on the CIFAR10.1 dataset, where the surrogate model achieves 87.90\% accuracy compared to the target model's 93.56\%, reflecting the impact of the distribution shift between CIFAR10 and CIFAR10.1. Despite this, on datasets like MNIST and SVHN, our method consistently produces surrogate models with accuracy ratios close to or exceeding 99\%, as seen with the Resnet-18 model on MNIST achieving a near-perfect ratio of 99.99\%. These findings underscore the robustness of our approach in generating high-quality surrogate models across various datasets, even in the presence of distribution shifts.

\subsection{Universarlity Evaluation}

\textbf{Noise selection.}
We optimize the hyperparameters of the noise PDF using a grid search. In the experiment, we use a noise distribution that follows $\propto e^{-|x / \alpha|^\beta}$, where we optimize $\beta$ and set $\alpha$ such that $\sigma = 1$. This choice of $\sigma = 1$ helps achieve a larger certified radius and good performance across different certified radii, as shown in previous work~\cite{cohen2019certified}. The grid search is conducted on the SVHN dataset, where a model is trained and evaluated for each round of search. For each pair of parameters $\alpha$ and $\beta$, we train a ResNet-18 model with randomized smoothing. Finally, as an approximation to Eq.~\eqref{eq:score}, we compute the robust score on a set of $\sigma$ = [0.25, 0.50, 0.75, 1.00] to evaluate the overall performance. To obtain the certified radii, we use Monte Carlo sampling of size 1000.

\begin{table*}[h!]
  \caption{Robust score under different noise distributions.}
  \resizebox{\linewidth}{!}{%
    \renewcommand{\arraystretch}{1.4} 
    \begin{tabular}{|c|cccccccccccccccc|}
      \hline
      $\beta$ &0.25 &0.50 &0.75 &1.00 &1.25 &1.50 &1.75 &2.00 &2.25 &2.50 &2.75 &3.00 &3.25 &3.50 &3.75 &4.00  \\
      \hline
      $\ell_1$ &1.212 &1.854 &\textbf{1.994} &1.741 &1.728 &1.701 &1.687 &1.622 &1.660 &1.581 &1.569 &1.511 &1.499 &1.472 &1.462 &1.459 \\
\hline
      $\ell_2$ &0.000 &0.001 &0.722 &0.894 &0.941 &1.156 &1.223 &1.424 &1.548 &1.644 &\textbf{1.708} & 1.633 &1.588 &1.490 &1.332 &1.264 \\
\hline     
      $\ell_\infty$ &0.000 &0.008&0.023 &0.041 &0.053&0.064&0.068&0.071 &0.073&0.075&\textbf{0.079} & 0.076 &0.074 &0.063 &0.060 &0.056 \\
      \hline
    \end{tabular}%
    }
  \label{tab:Robust score under different noise distributions}
\end{table*}
As indicated by the results in Table \ref{tab:Robust score under different noise distributions}, the optimal $\beta$ for $\ell_1$ norm is 0.75, while that for $\ell_2$ and $\ell_\infty$ norms is 2.75. This suggests that the commonly used Laplacian noise with $\beta = 1$ and Gaussian noise with $\beta = 2$ are not optimal for certified defense across $\ell_1$, $\ell_2$, and $\ell_\infty$ norms. A smaller $\beta$ strikes a better balance between the certified radius and certified accuracy. The results show that the choice of noise distributions is crucial for the performance of the certified defense.

\begin{figure*}[!t]
    \centering
    \subfloat[$\ell_1$]{\includegraphics[width=0.31\linewidth]{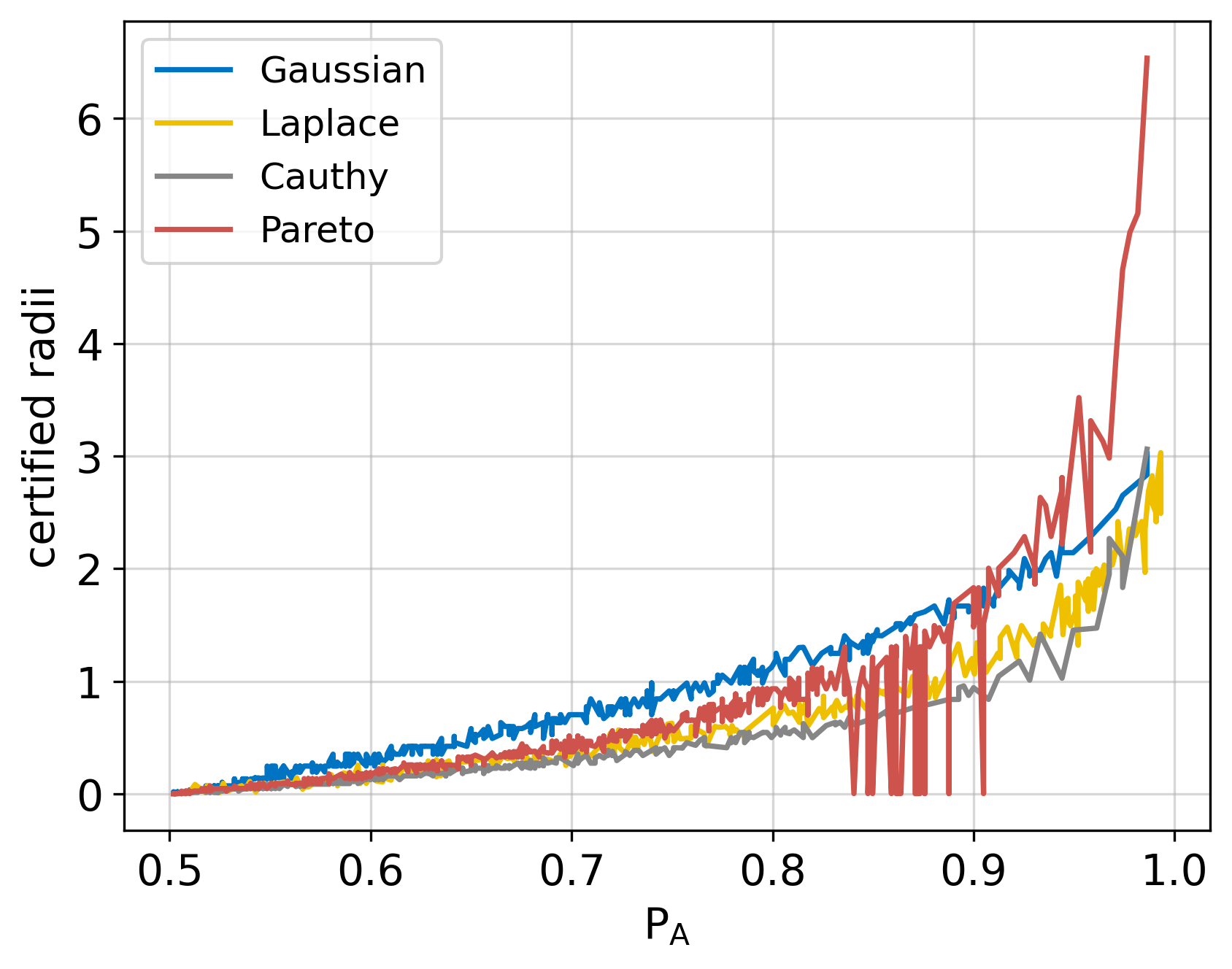} } \hspace{1mm}
    \subfloat[$\ell_2$]{\includegraphics[width=0.31\linewidth]{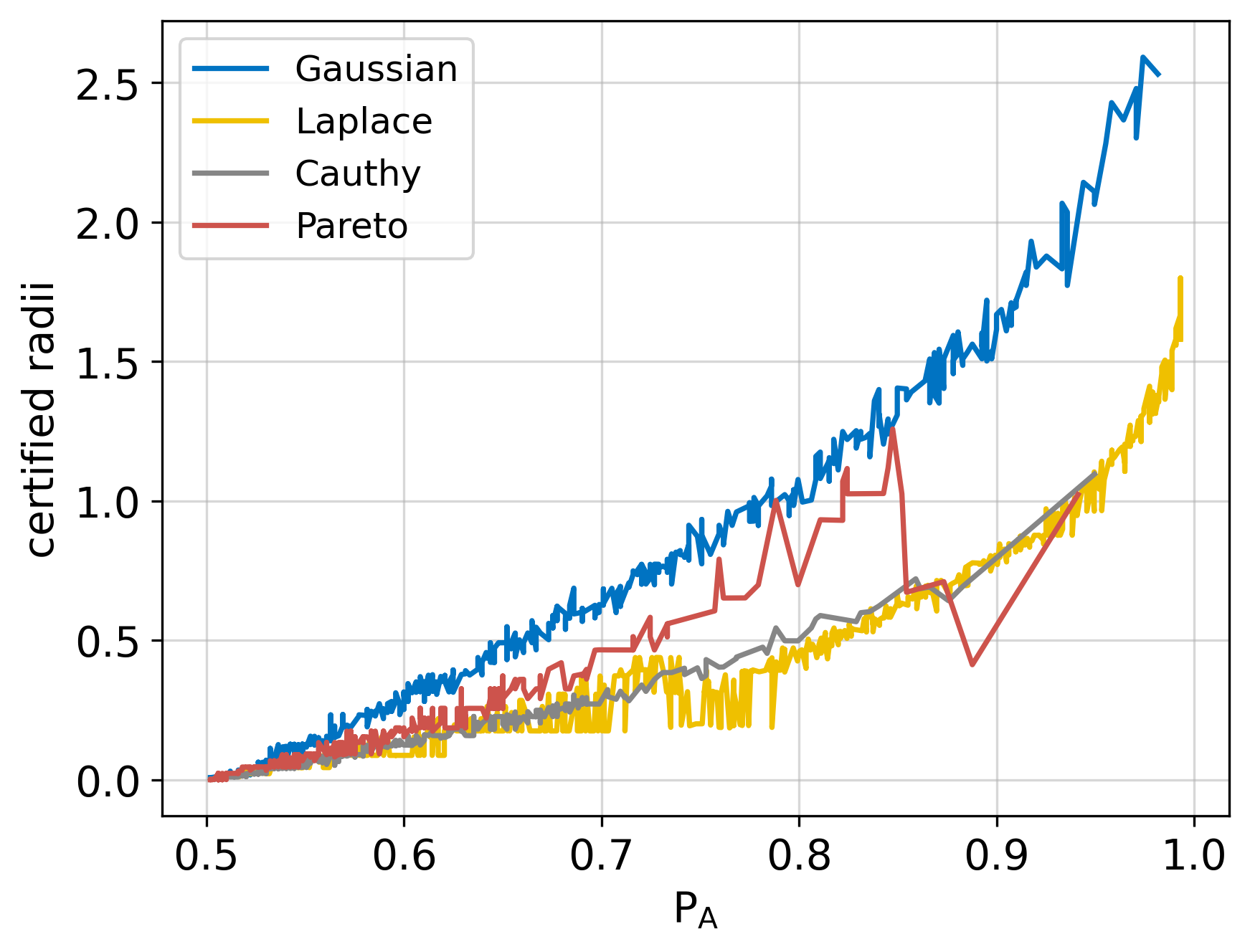} }\hspace{1mm}
    \subfloat[$\ell_\infty$]{\includegraphics[width=0.31\linewidth]{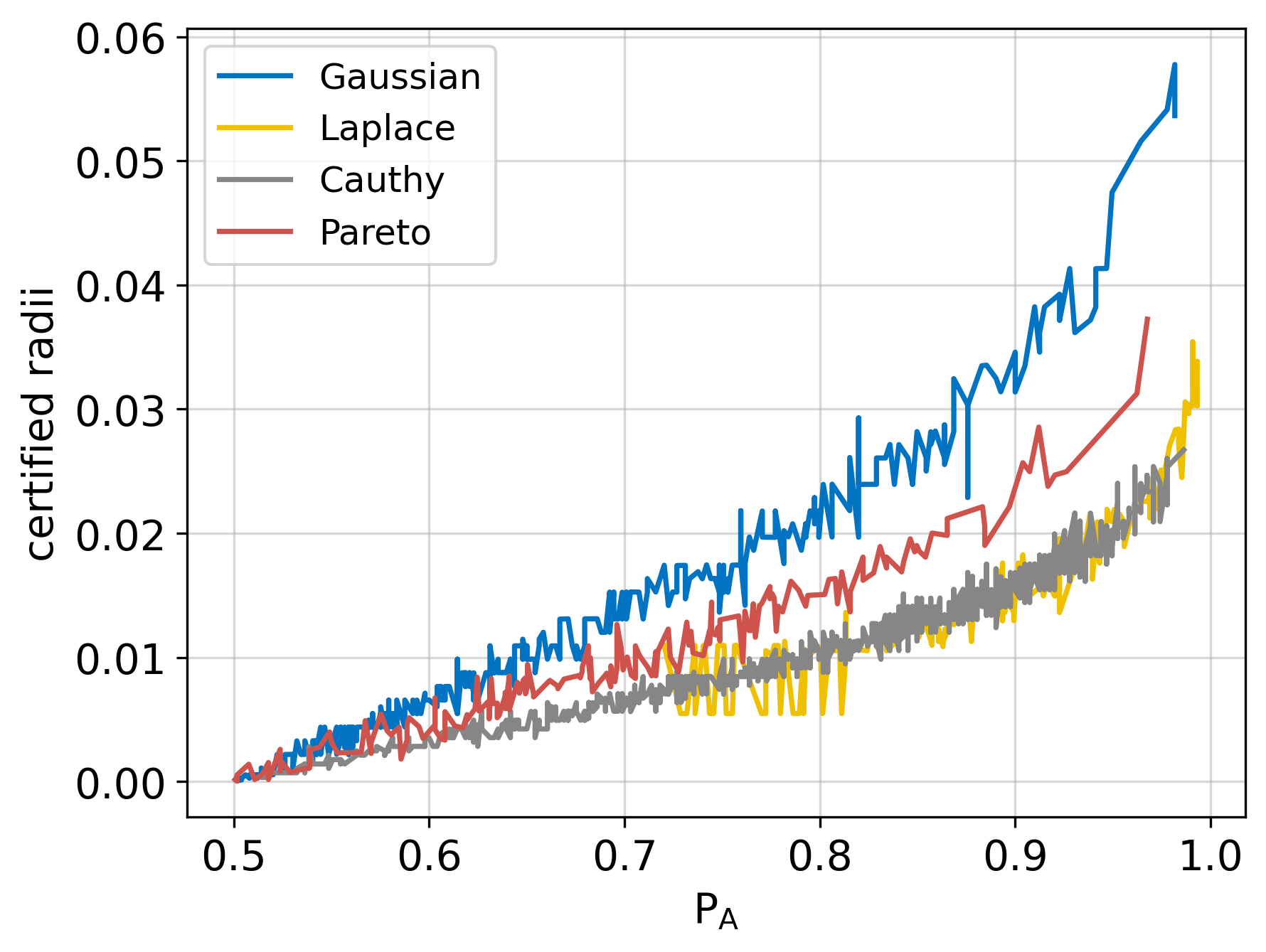} }
    \caption{$p_A$-$R$ curves for different noise distributions.}
    \label{fig:p-r-different-noise}
\end{figure*}

\begin{figure*}[!t]
    \centering
    \subfloat[$\ell_1$]{\includegraphics[width=0.31\linewidth]{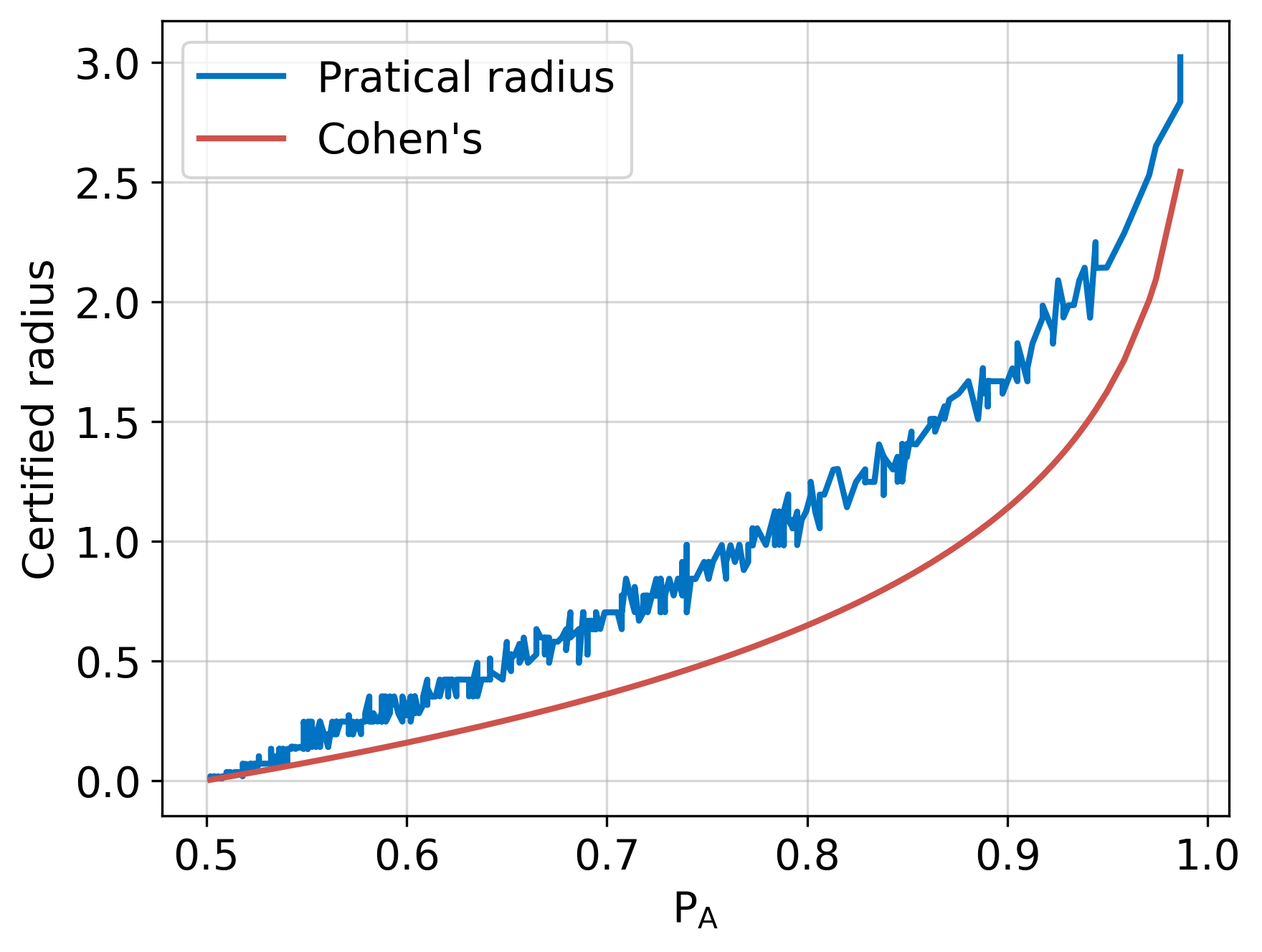} } \hspace{1mm}
    \subfloat[$\ell_2$]{\includegraphics[width=0.31\linewidth]{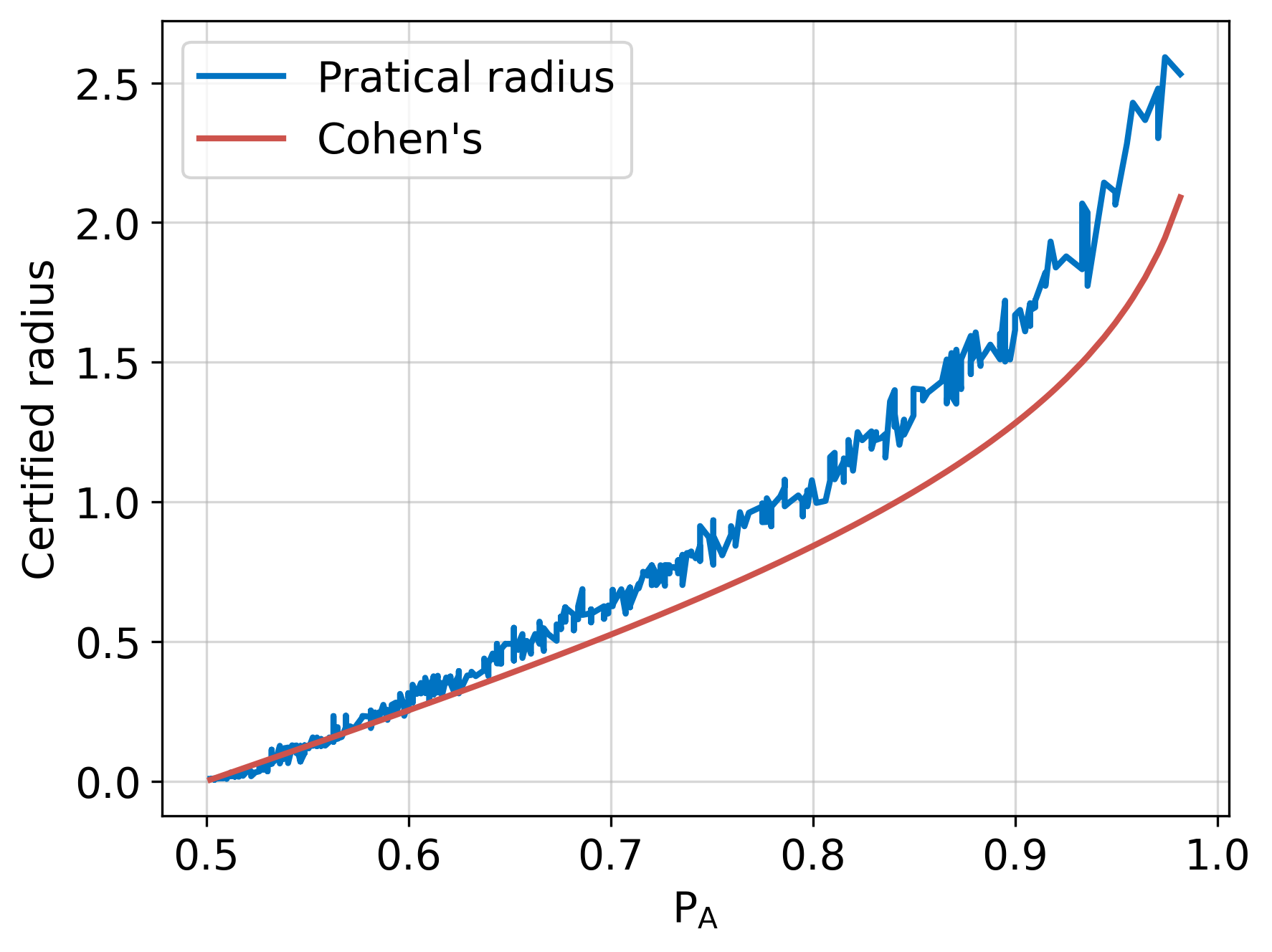} }\hspace{1mm}
    \subfloat[$\ell_\infty$]{\includegraphics[width=0.31\linewidth]{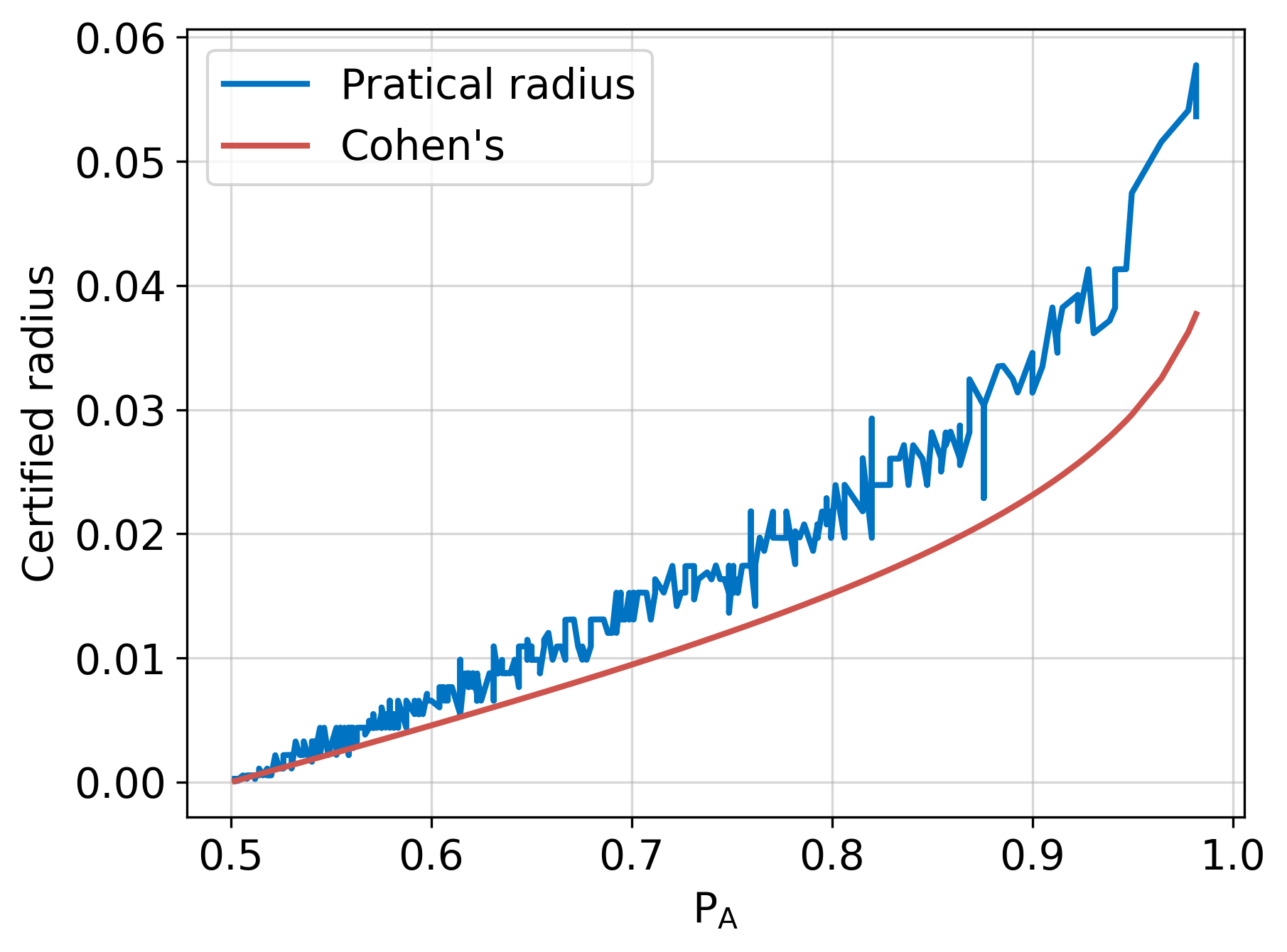} }
    \caption{$p_A$-$R$ curve comparisons between DUCD and RS~\cite{cohen2019certified}.}
    \label{fig:p-r-DUCD-and-cohen}
\end{figure*}


\begin{table*}[hbt]
  \caption{Certified accuracy under different certified radius and norms.}
  \resizebox{\linewidth}{!}{%
  \renewcommand{\arraystretch}{1.4}
    \begin{tabular}{|c|ccccccccccccc|}
      \hline
      $\ell_1$-radius  & 0.25 &0.50 &0.75 & 1.00 &1.25 & 1.50 &1.75&  2.00&2.25 & 2.5&2.75&3.00 & score \\
      \hline
      RS ~\cite{cohen2019certified}&72.93&60.47&54.37&48.86&41.73&36.75&35.23&30.77&30.82&26.75&26.60&25.27&1.103\\
      WUCD&\textbf{77.37}&65.06&\textbf{55.85}&48.93&43.17&\textbf{40.02}&36.14&\textbf{39.90}&\textbf{31.56}&\textbf{30.92}&28.36&25.79&\textbf{1.179}\\
    DUCD &76.68&\textbf{65.33}&55.65&\textbf{49.50}&\textbf{43.85}&39.90&\textbf{37.44}&33.74&31.16&29.28&27.67&25.54&1.161\\
    DUCD (C10-S) &74.89&62.26&55.41&49.47&43.93&39.68&36.57&\textbf{34.10}&31.64&29.85&\textbf{28.87}&\textbf{27.16}&1.157\\
    DUCD (C10.1)&74.64&65.30&55.49&47.62&38.84&34.19&32.43&30.58&28.89&26.60&25.79&24.95&1.089\\
    \hline
    $\ell_2$-radius   & 0.25 &0.50 &0.75 & 1.00 &1.25 & 1.50 &1.75&  2.00&2.25 & 2.5&2.75&3.00 & score \\
    \hline
    ZO-AE-DS~\cite{zhangrobustify} & 12.92&12.65&12.33&11.68&11.55&11.16&11.13&11.09&11.06&11.15&10.66&10.55&0.315\\
    RS~\cite{cohen2019certified}&75.41&65.20&\textbf{57.11}&45.40&43.01&38.22&36.45&33.52&31.16&28.96&27.81&26.17&1.144\\
    WUCD  &\textbf{77.52}&64.94&55.89&\textbf{49.39}&42.87&\textbf{40.16}&35.92&\textbf{34.10}&30.94&29.97&27.43&26.07&1.159
 \\
    DUCD &76.00&64.28&55.88&49.15&43.50&39.95&\textbf{36.57}&33.20&\textbf{32.19}&\textbf{30.67}&\textbf{27.93}&26.71&\textbf{1.162}\\
    DUCD (C10-S) &74.63&63.29&55.49&49.18&\textbf{43.79}&40.11&36.36&33.86&31.33&29.93&28.42&\textbf{27.25}&1.157\\
    DUCD (C10.1) &76.54&\textbf{65.24}&55.21&46.92&38.81&34.45&31.84&29.54&27.71&26.89&25.43&24.77&1.082\\
    \hline
    $\ell_\infty$-radius & 0.25 &0.50 &0.75 & 1.00 &1.25 & 1.50 &1.75&  2.00&2.25 & 2.5&2.75&3.00 & score \\
    \hline
    RS~\cite{cohen2019certified}&77.02&64.38&56.51&48.29&\textbf{44.86}&38.75&35.89&32.27&30.06&\textbf{29.84}&\textbf{28.77}&27.08&1.154

 \\
    WUCD &\textbf{77.03}&65.09&\textbf{55.62}&48.89&42.97&\textbf{39.75}&35.57&34.10&30.65&29.77&27.52&26.06&1.153
 \\
    DUCD &76.11&64.54&55.26&49.20&44.21&39.66&36.61&34.01&31.09&30.01&26.50&26.42&\textbf{1.156}
 \\
    DUCD (C10-S) &76.04&65.30&54.83&\textbf{49.32}&43.77&39.74&\textbf{37.24}&33.74&\textbf{31.55}&29.77&28.07&\textbf{27.19}&1.162
 \\
    DUCD (C10.1) &76.22&\textbf{65.62}&\textbf{55.62}&47.24&38.58&34.43&32.11&29.77&28.13&26.77&25.47&24.85&1.086
 \\
      \hline
    \end{tabular}%
  }
  \label{tab:Certified accuracy under different certified radius and norm}
\end{table*}

\textbf{Certified radius.}
In theory, randomized smoothing provides certified robustness against input perturbations under any $\ell_p$-norm constraints. Consequently, our experimental evaluation focuses on assessing the certified radii under different noise probability density functions (PDFs) for $p=1$, $p=2$, and $p=\infty$. To identify the optimal PDF for each $\ell_p$ perturbation, we compute the certified radii for various PDFs, all with the same variance, across a range of $p_A$ values within the interval (0.5, 1.0].

As illustrated in Fig. \ref{fig:p-r-different-noise}, the $p_A$-R curves for Gaussian, Laplacian, Pareto, and Cauchy distributions show that, for all $\ell_p$ norms and most $p_A$ values, Gaussian noise generally yields the largest certified radii. Except for the Cauchy distribution, the curves for the other noise distributions are quite similar. However, for lower $p_A$ values and $\ell_2$ and $\ell_\infty$ perturbations, deriving a certified radius using Laplace noise proves challenging, which aligns with findings from previous research~\cite{yang2020randomized}. This difficulty underscores the limitations of Laplace noise in achieving robust certification under certain conditions.

\subsection{Defense Performance}
We evaluate DUCD's defense performance against existing methods, including a black-box defense ZO-AE-DS~\cite{zhangrobustify}.
Note that ZO-AE-DS only provides a certified defense under $\ell_2$-norm constraints, hence it is not evaluated for $\ell_1$ and $\ell_\infty$ norms. In addition, as a strong baseline, we also treat the original model as a white-box model, to which we can apply white-box defense methods including randomized smoothing (RS)~\cite{cohen2019certified}. As an ablation, we also apply the same defense used by DUCD to the original target model and obtain a model with white-box universal certified defense (WUCD).
For black-box scenarios, the query budget for both DUCD and the baseline methods is 20M.

Robustness scores are computed with $\sigma$ =[0.25, 0.5, 0.75, 1.0]. Laplace noise is used
for $\ell_1$ defense, and Gaussian noise is used for $\ell_2$ and $\ell_\infty$ defense. For both our method and the baselines, the Monte Carlo sampling size is set to 4000. The certified accuracies are obtained with the certified radii of 500 random samples from the test set.

In Table \ref{tab:Certified accuracy under different certified radius and norm}, it is evident that DUCD significantly outperforms the baseline black-box defense, ZO-AE-DS, in both certified accuracy and robustness score. Specifically, DUCD shows a remarkable improvement in robustness score for the $\ell_2$-norm, with an increase of 0.845 compared to ZO-AE-DS.
When compared to Cohen's white-box method, although the NBCD approach exhibits a slightly lower certified accuracy, its robustness score remains highly competitive, showing increases of 6.33\%, 3.50\%, and 1.81\% under the $\ell_1$, $\ell_2$, and $\ell_\infty$ norms, respectively. These results indicate that NBCD can more effectively resist adversarial attacks, thereby enhancing the overall robustness of the model.

Furthermore, as shown in Fig.~\ref{fig:p-r-DUCD-and-cohen}, we compare the robust radius across all $\ell_p$ norms with Cohen’s randomized smoothing (RS) method and plot the $p_A$-R curves. The certified radius achieved by DUCD consistently surpasses the results obtained by the RS method, demonstrating that our approach provides superior robust certification performance for input data with any continuous noise distribution across all $\ell_p$ norms. This highlights the versatility and effectiveness of DUCD in maintaining robustness under diverse conditions.

\subsection{Adaptive Attacks}

%
We empirically evaluate the defense performance of different defense methods on CIFAR10 against popular adversarial attacks, including AutoPGD~\cite{croce2020reliable}, ACG~\cite{yamamura2022diversified}, HSJA~\cite{chen2020hopskipjumpattack}, and Square Attack~\cite{andriushchenko2020square}. Among them, AutoPGD and ACG are white-box attacks, whereas HSJA and Square Attack are black-box attacks.

\begin{figure*}[!t]
    \centering
    \subfloat[$\ell_1$ - AutoPGD]{\includegraphics[width=0.31\linewidth]{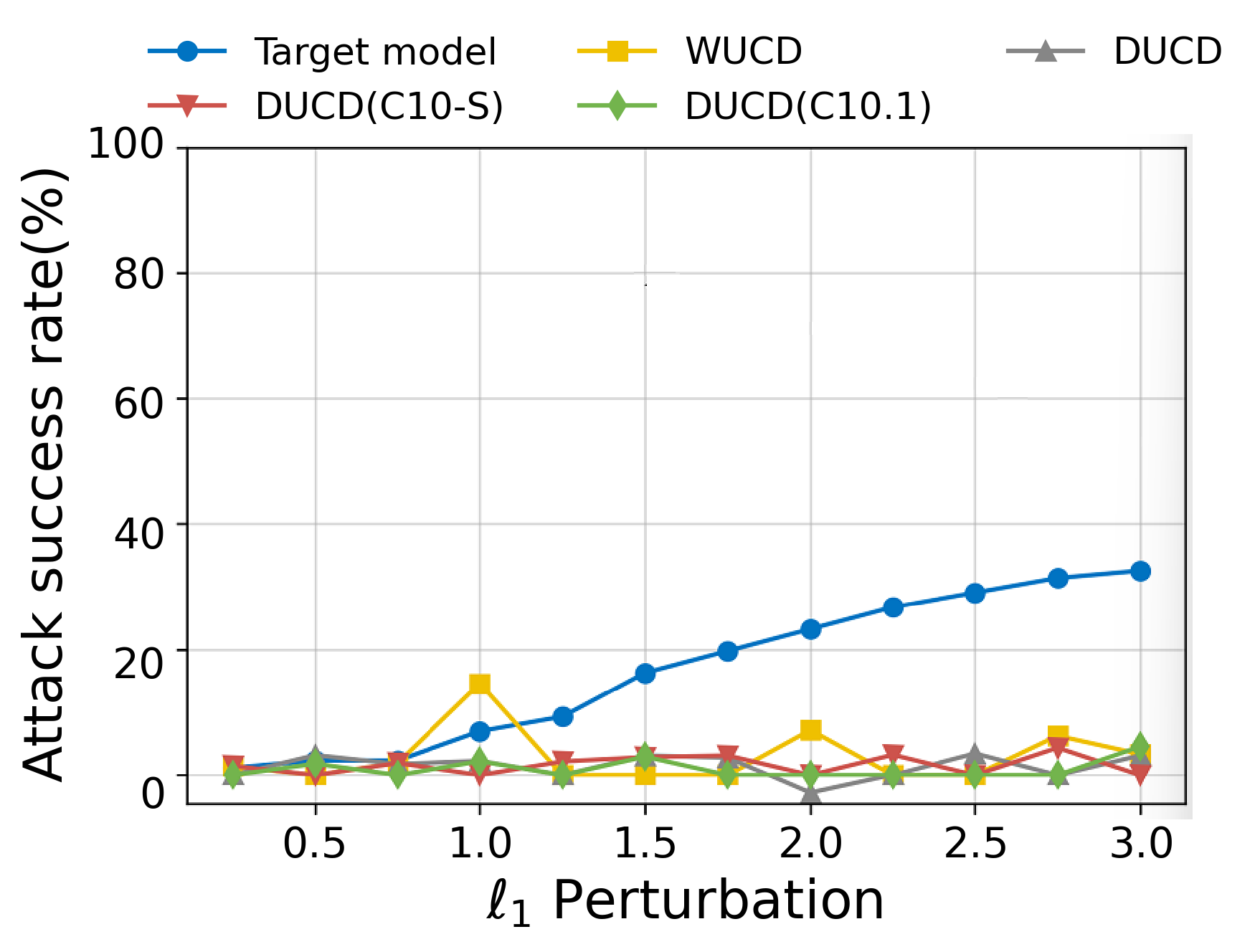} } \hspace{1mm}
    \subfloat[$\ell_2$ - AutoPGD]{\includegraphics[width=0.31\linewidth]{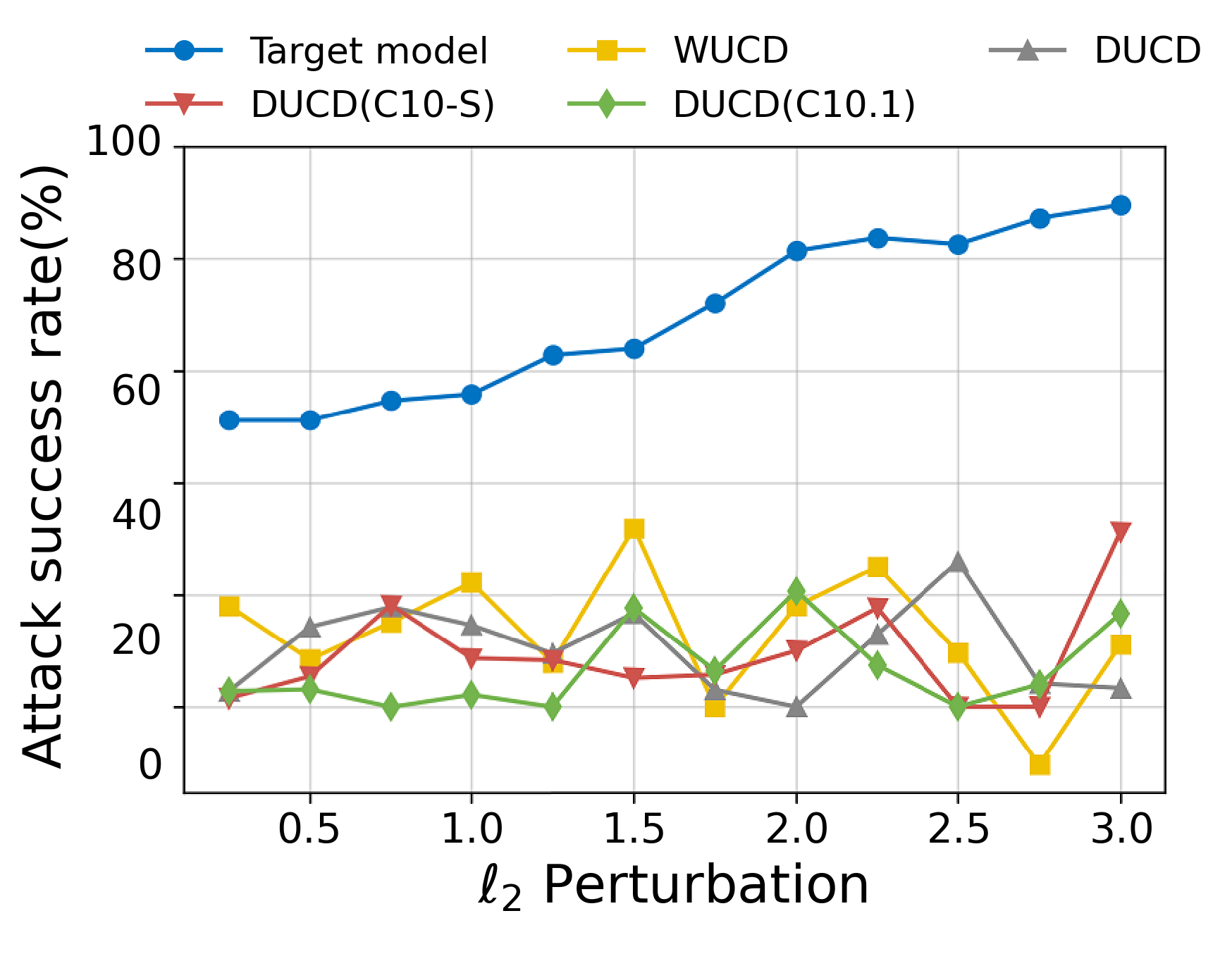} }\hspace{1mm}
    \subfloat[$\ell_\infty$ - AutoPGD]{\includegraphics[width=0.31\linewidth]{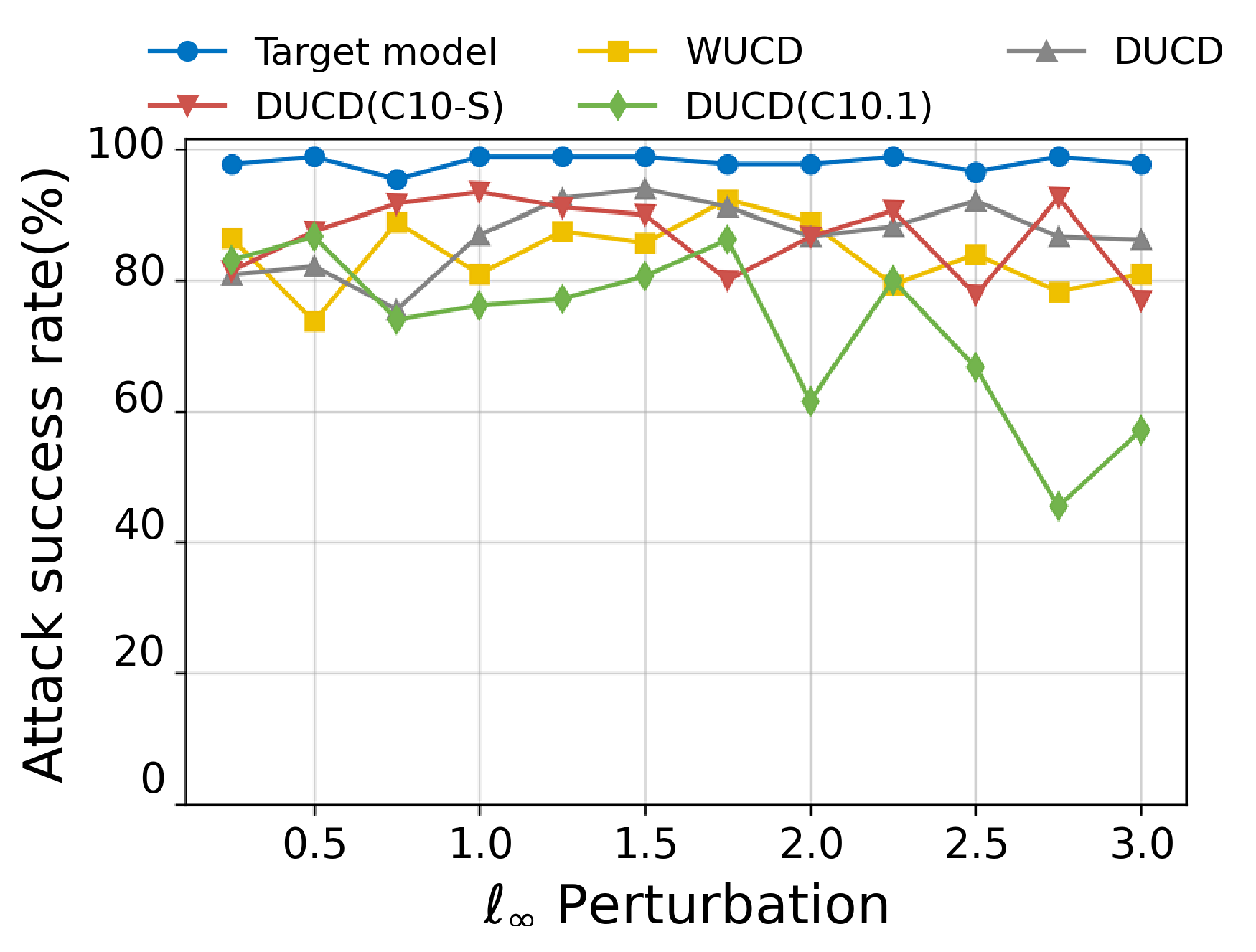} }\\
    \subfloat[$\ell_1$ - ACG]{\includegraphics[width=0.31\linewidth]{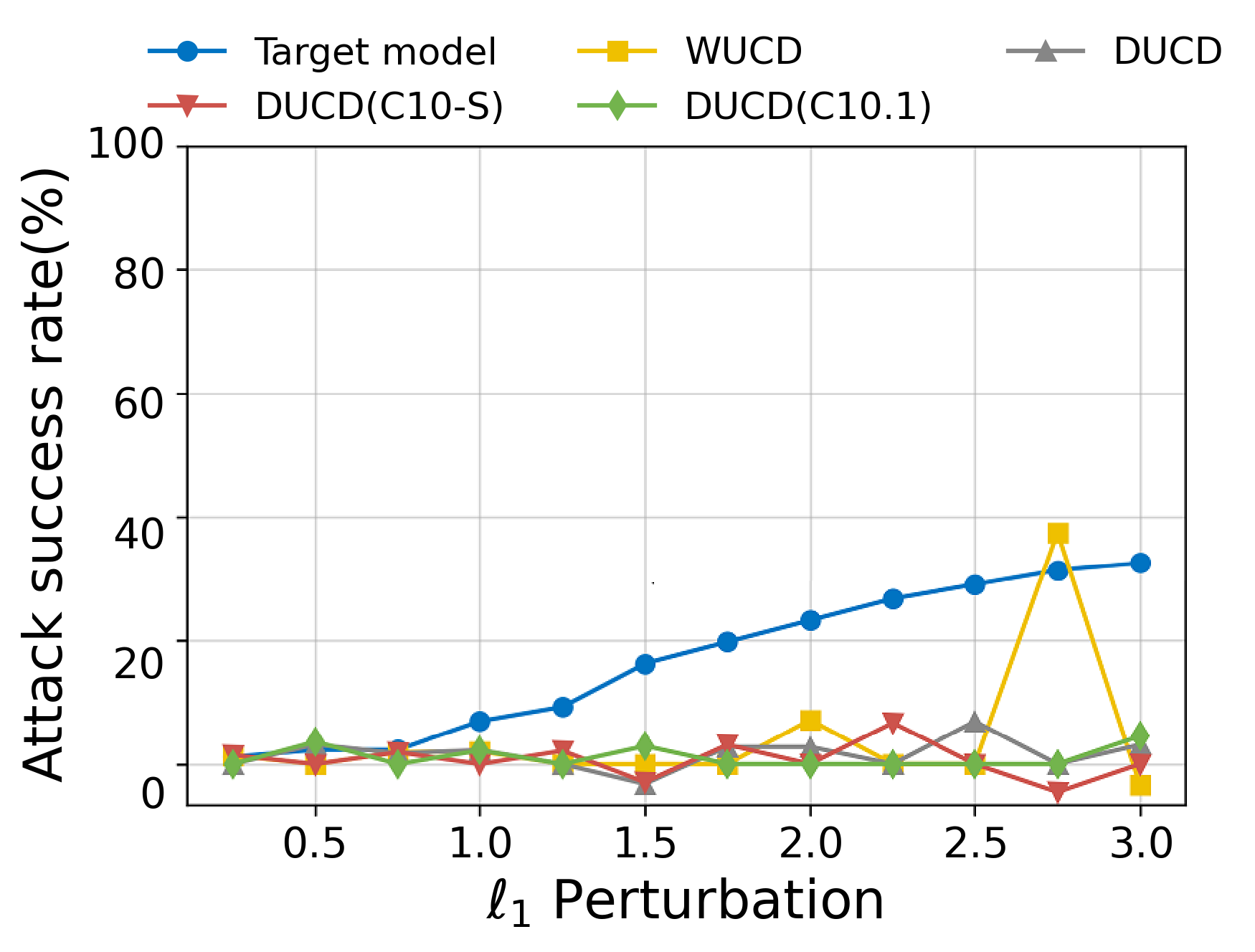} } \hspace{1mm}
    \subfloat[$\ell_2$ - ACG]{\includegraphics[width=0.31\linewidth]{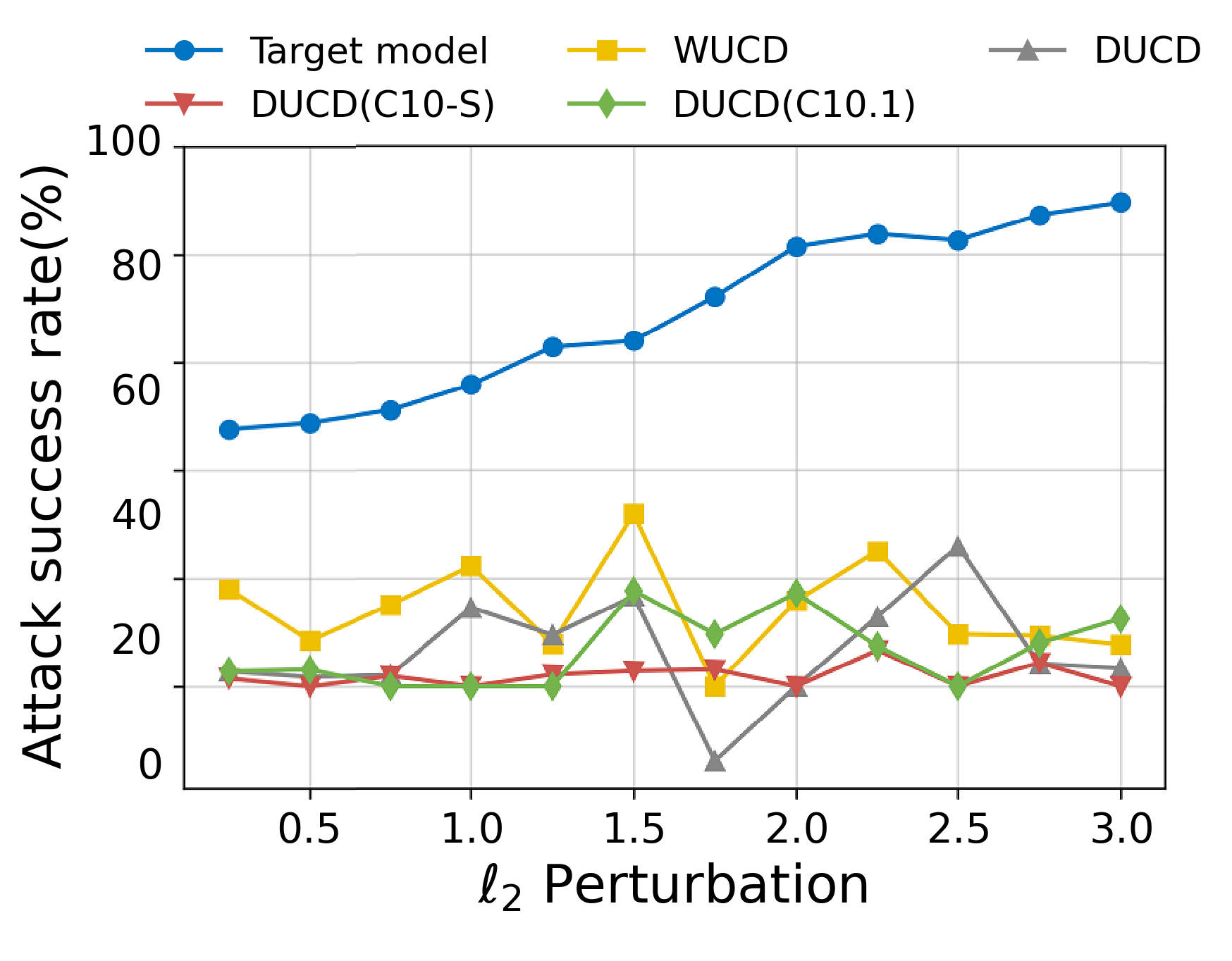} }\hspace{1mm}
    \subfloat[$\ell_\infty$ - ACG]{\includegraphics[width=0.31\linewidth]{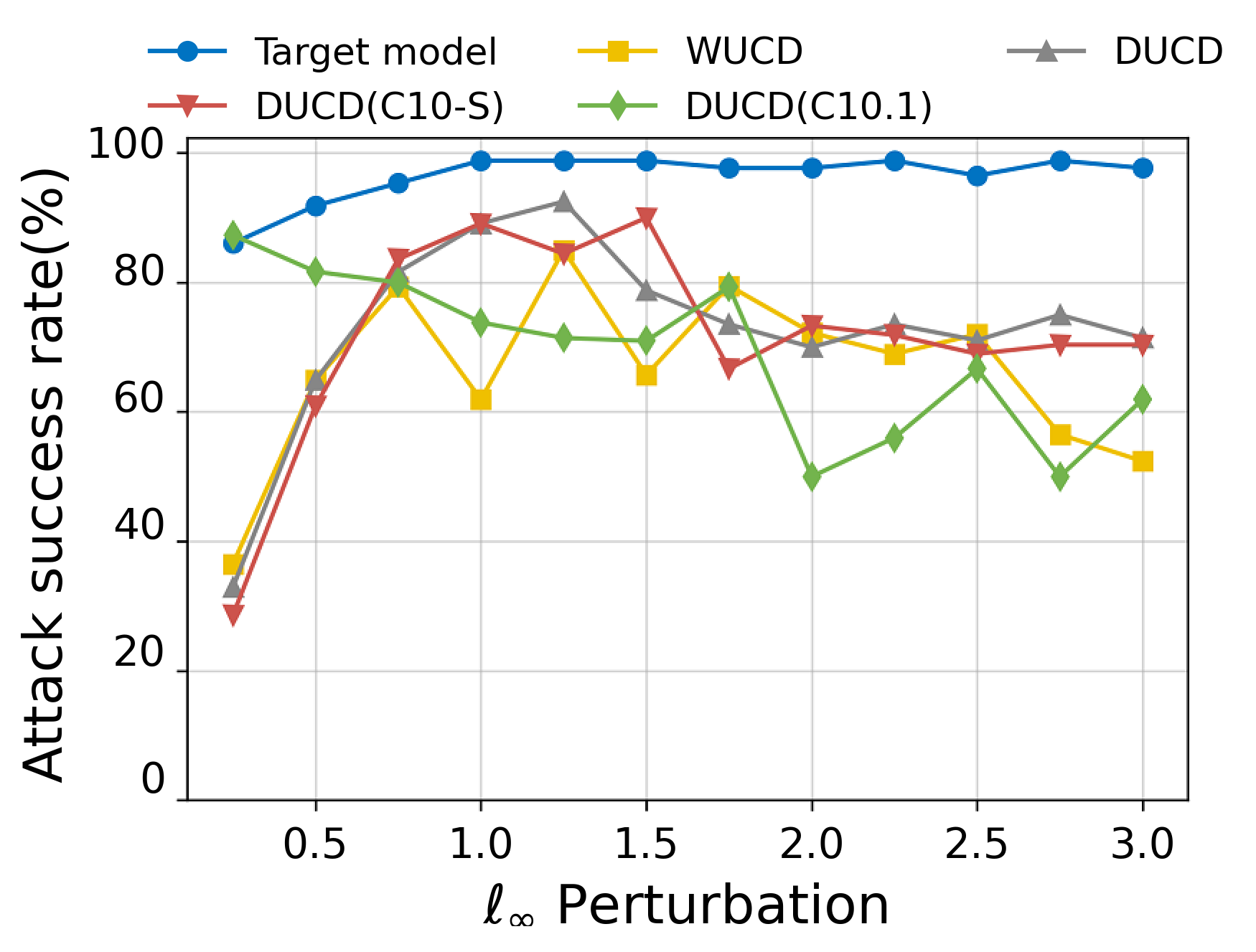} }\\
    \caption{Defense performance under different white-box adversarial attacks, including AutoPGD (a), (b), (c) and ACG (d), (e), (f)}
    \label{fig:white-box}
\end{figure*}

\begin{figure*}[!t]
    \centering
    \subfloat[$\ell_2$ - HSJA]{\includegraphics[width=0.23\linewidth]{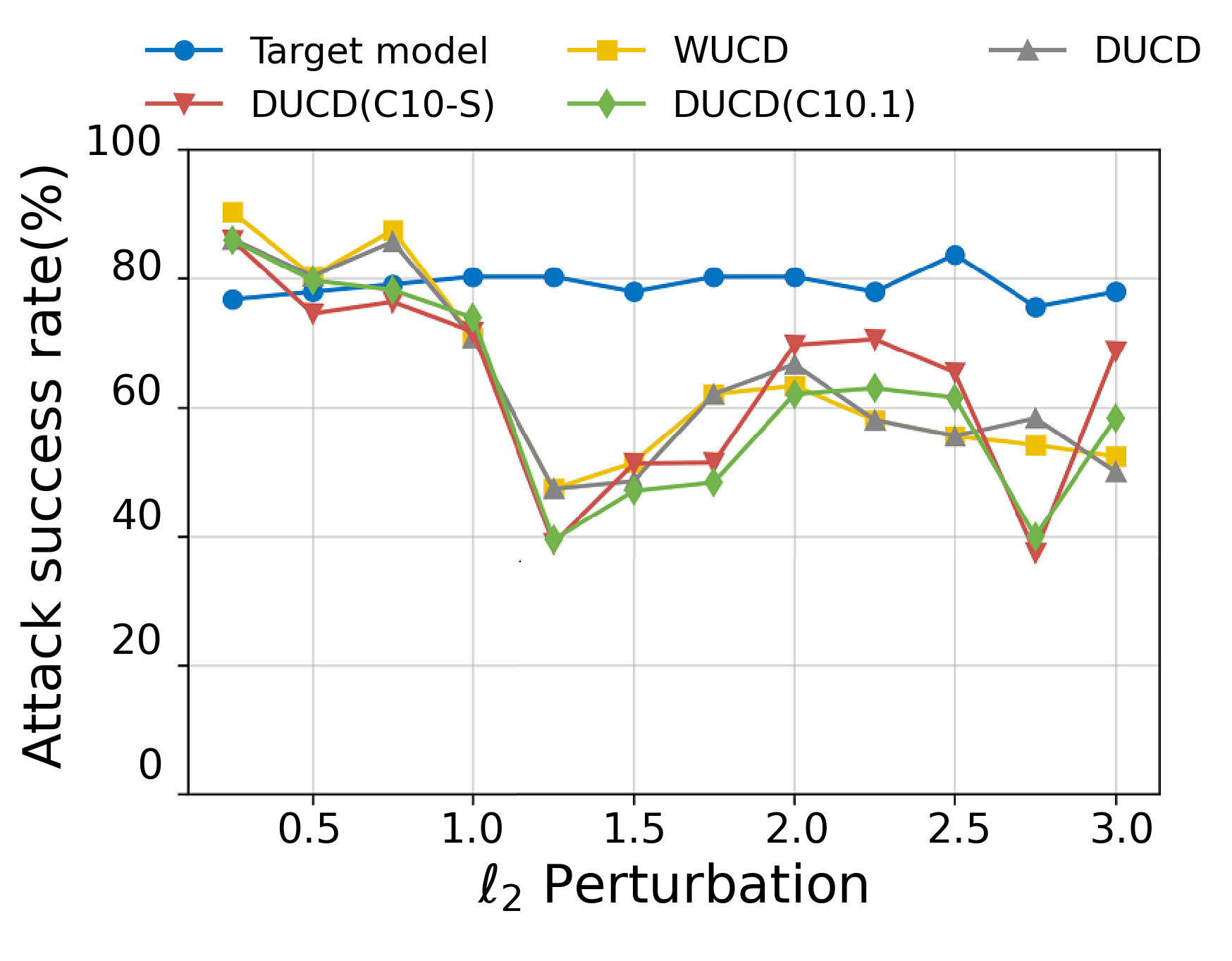} } \hspace{1mm}
    \subfloat[$\ell_\infty$ - HSJA]{\includegraphics[width=0.23\linewidth]{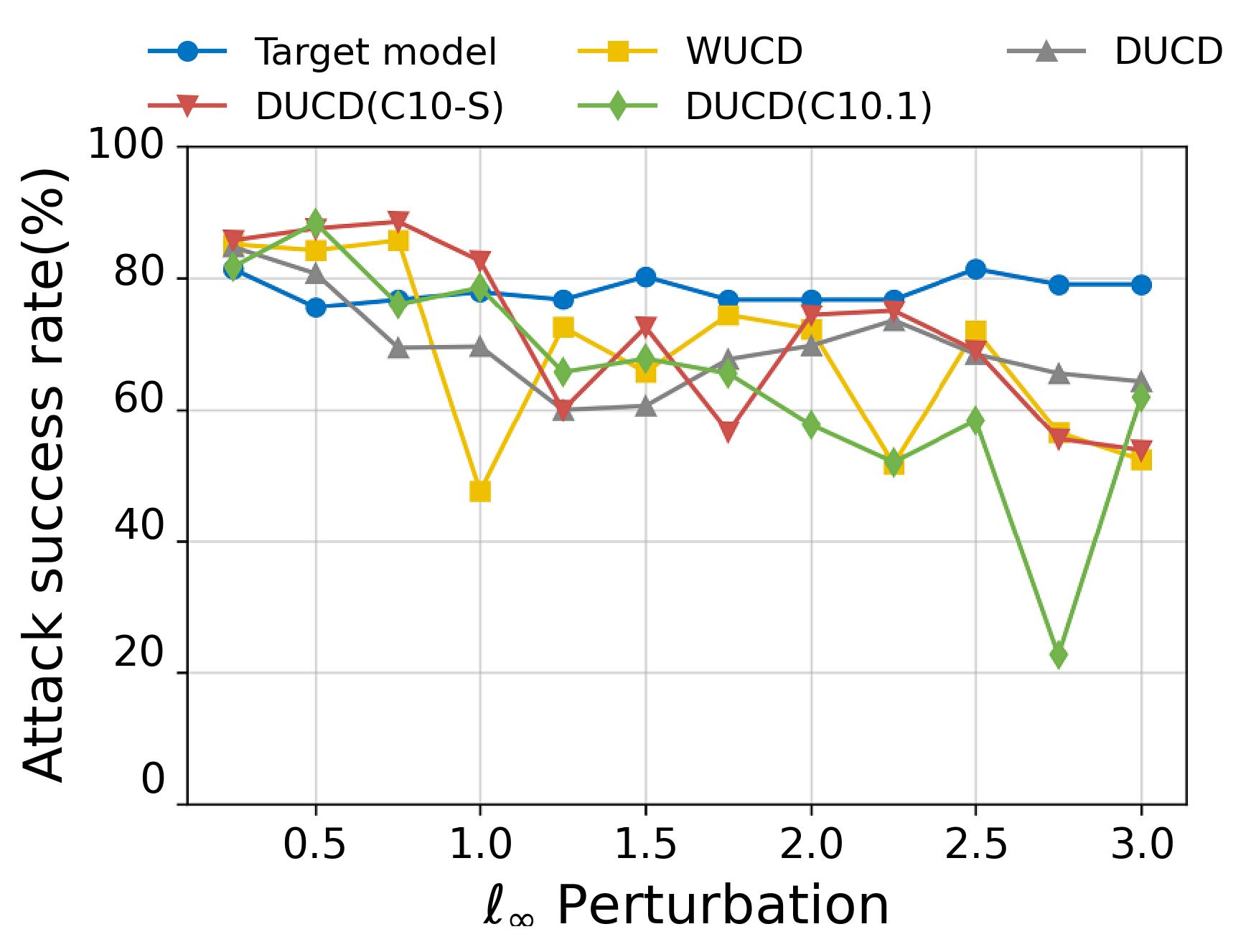} } \hspace{1mm}
    \subfloat[$\ell_2$ - Square attack]{\includegraphics[width=0.23\linewidth]{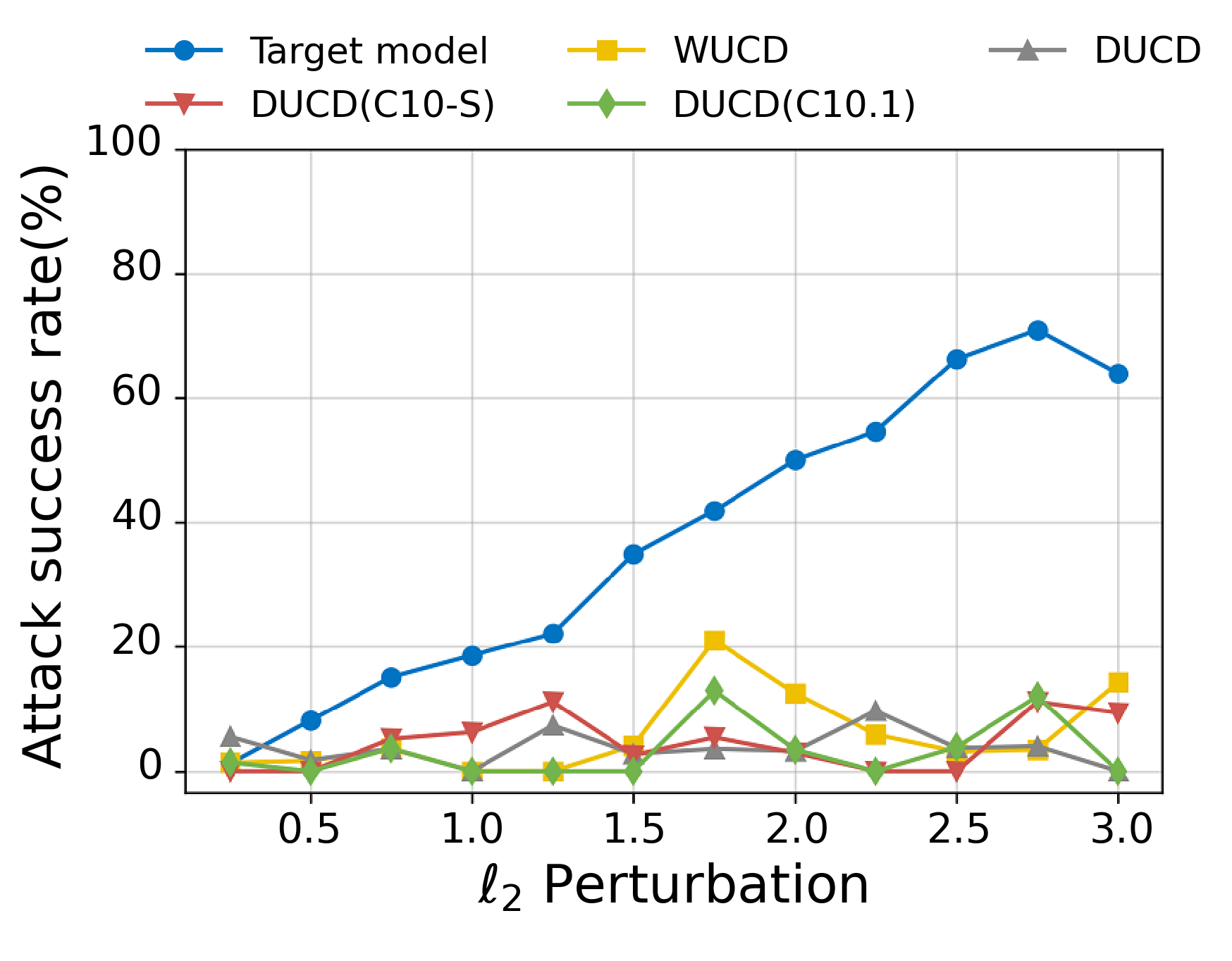} } \hspace{1mm}
    \subfloat[$\ell_\infty$ - Square attack]{\includegraphics[width=0.23\linewidth]{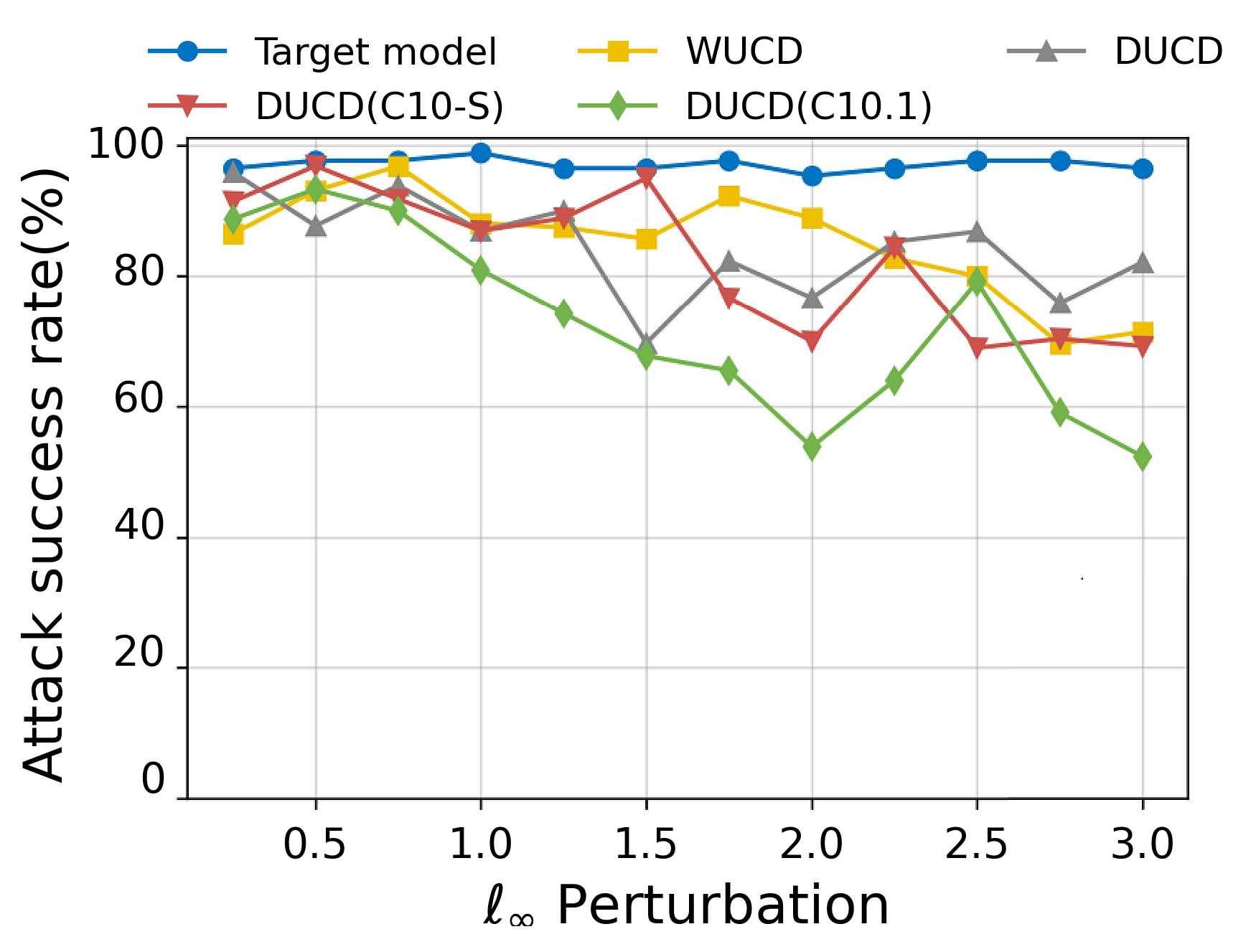} }\\
    \caption{Defense performance under different black-box adversarial attacks.}
    \label{fig:black-box}
\end{figure*}

In Fig. \ref{fig:white-box}, compared to white-box certified defense WUCD, DUCD consistently demonstrates competitive performance even when distillation is conducted on a different dataset (see DUCD (C10-S) and DUCD (C10.1)).
Under $\ell_1$ and $\ell_2$ norm attacks, as adversarial perturbations increase, the attack success rate (ASR) for white-box attacks rises to 90\%, while our defense method keeps the ASR below 20\%. 
However, for $\ell_\infty$ attacks, none of the defense schemes achieve satisfactory performance.
Under $\ell_\infty$ norm attacks, the ASR remains close to 100\%. In this case, the defense performance of DUCD and WUCD are similar, reducing the ASR by nearly 10\%. Notably, DUCD (C10.1) reduces the ASR to below 50\%, which may be due to the different distributions between the surrogate model's dataset and the original model's dataset, leading to a decrease in the effectiveness of the attack method when rely on one dataset.
In Fig. \ref{fig:black-box}, we showcase defense performance against query-based black-box attacks, concentrating our analysis on $\ell_2$ and $\ell_\infty$ attack scenarios. 
This focus is a result of the inherent limitation of these two query-based attack methods, which exclusively target the $\ell_2$ and $\ell_\infty$ norms.
Similar to the white-box attack, DUCD performs similarly to WUCD in practice.

However, when subjected to $\ell_\infty$ norm attacks, all defense methods, including our proposed approach, demonstrate suboptimal performance. This shortfall is particularly evident even in the WUCD method, which, despite having rigorous theoretical guarantees as outlined in prior research~\cite{hong2022unicr}, fails to achieve robust effectiveness in practice. The underlying reason for this limitation is closely tied to the nature of the certification process itself. Certified defenses, such as those described by Cohen et al.~\cite{cohen2019certified}, rely on a stringent certification criterion, whereby only a subset of input samples—specifically, those that successfully pass the certification—can be guaranteed to resist adversarial perturbations within a defined radius. In the case of $\ell_\infty$ norm attacks, which involve highly localized and precise perturbations, the certification process often struggles to verify the robustness of the inputs. Consequently, the defense mechanisms become less effective, as a significant portion of the input data fails to meet the necessary criteria for certification. This challenge highlights a critical area for further research, particularly in developing more resilient defenses that can reliably certify robustness across a broader spectrum of input data, even under the more challenging $\ell_\infty$ norm constraints.

\subsection{Purification}
We introduce a purification experiment that employs a randomized smoothing certification process. This process filters out uncertified inputs, ensuring that only certified samples are forwarded to the classifier. As demonstrated in the study by~\cite{hong2022unicr}, certified samples exhibit
greater robustness, making them more resilient to minor noise. Through the randomized smoothing certification process, the target classifier is more likely to accurately classify certified samples.
\begin{figure}[!t]
\centering
\includegraphics[width=0.8\linewidth]{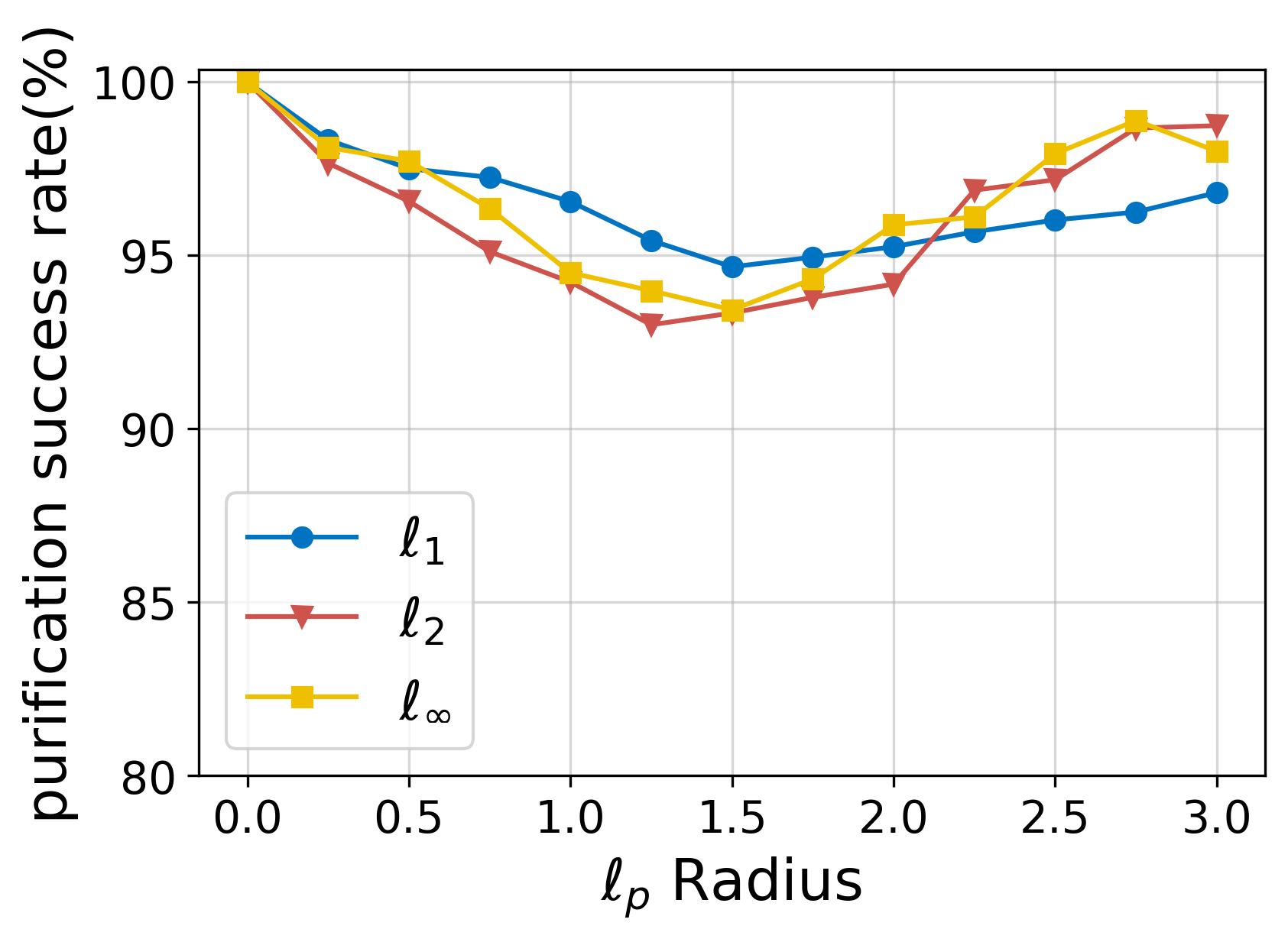} 
\caption{Purification success rate under different $\ell_p$ norm}
\label{fig:purification}
\end{figure} 

We conduct experiments using the CIFAR10 dataset with a fixed number of 1000 Monte Carlo samples. The purification success rate is defined as the percentage of certified samples out of all inputs.
We generated plots for various noise intensities, including under
$\ell_1$, $\ell_2$, and $\ell_\infty$ norm. The results are presented in Fig \ref{fig:purification}.


As shown in Fig \ref{fig:purification}, when the noise level is set to 0, the purification pass rate approaches 100\%.  Although there are some fluctuations in the success rate, the baseline rate remains above 90\%. This indicates that our method, as a preprocessing step, does not filter out clean samples Our purification experiments demonstrate that the DUCD method can be employed as a
pre-processing module to filter inputs before reaching the target
classifier. The pre-process also effectively enhances the robustness of the target classifier.

\subsection{Privacy Evaluation}
Finally, we conduct a membership inference attack~\cite{song2019privacy} to evaluate the privacy protection performance of the proposed surrogate model generation method. The goal of a membership inference attack is to determine whether a specific data point is in the training dataset of the target model~\cite{wu2024rethinking}, which can violate privacy guarantees, especially in scenarios where the training dataset contains sensitive or personal information.

\begin{figure*}[!t]
    \centering
    \subfloat[$\ell_1$]{\includegraphics[width=0.31\linewidth]{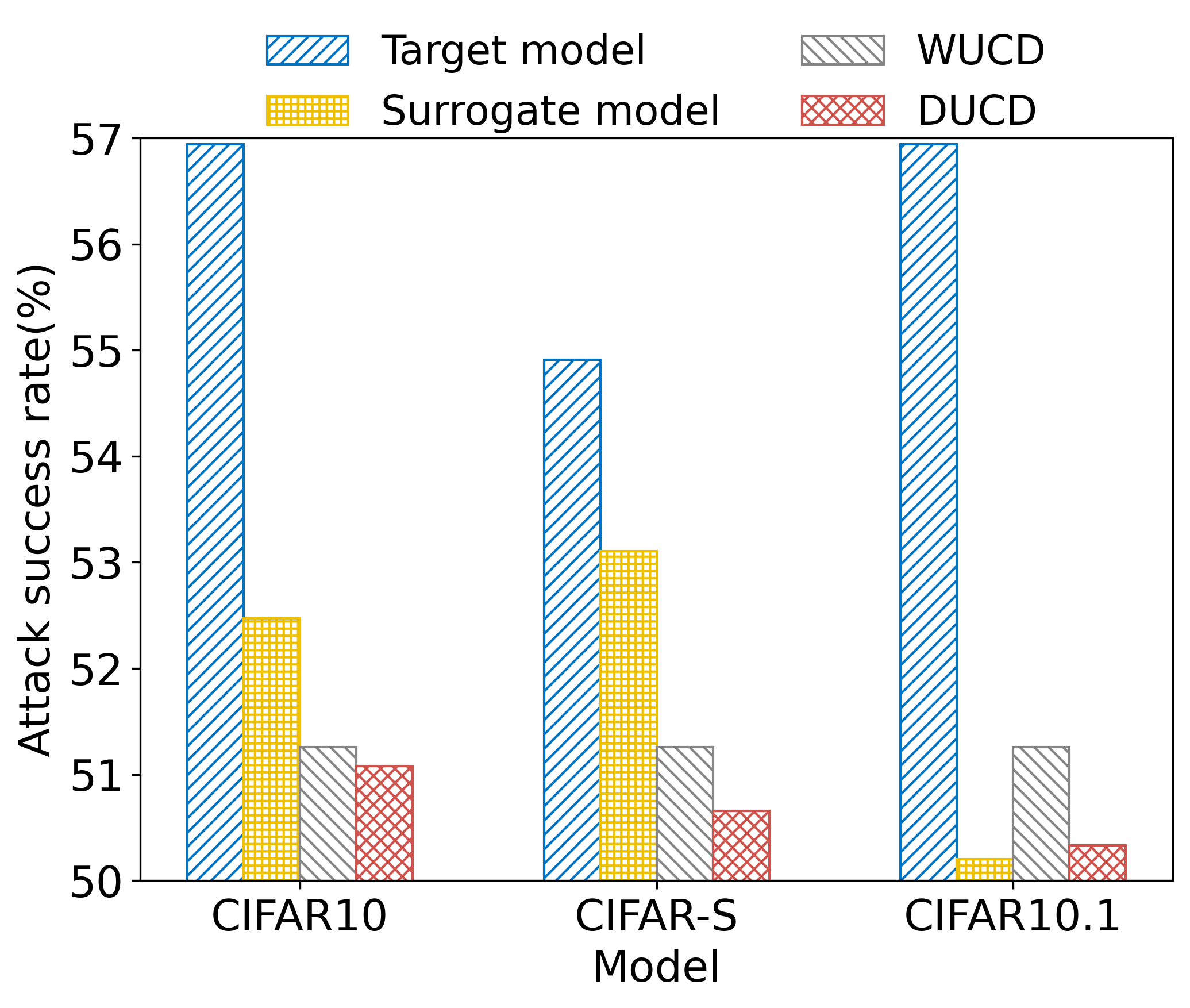} } \hspace{1mm}
    \subfloat[$\ell_2$]{\includegraphics[width=0.31\linewidth]{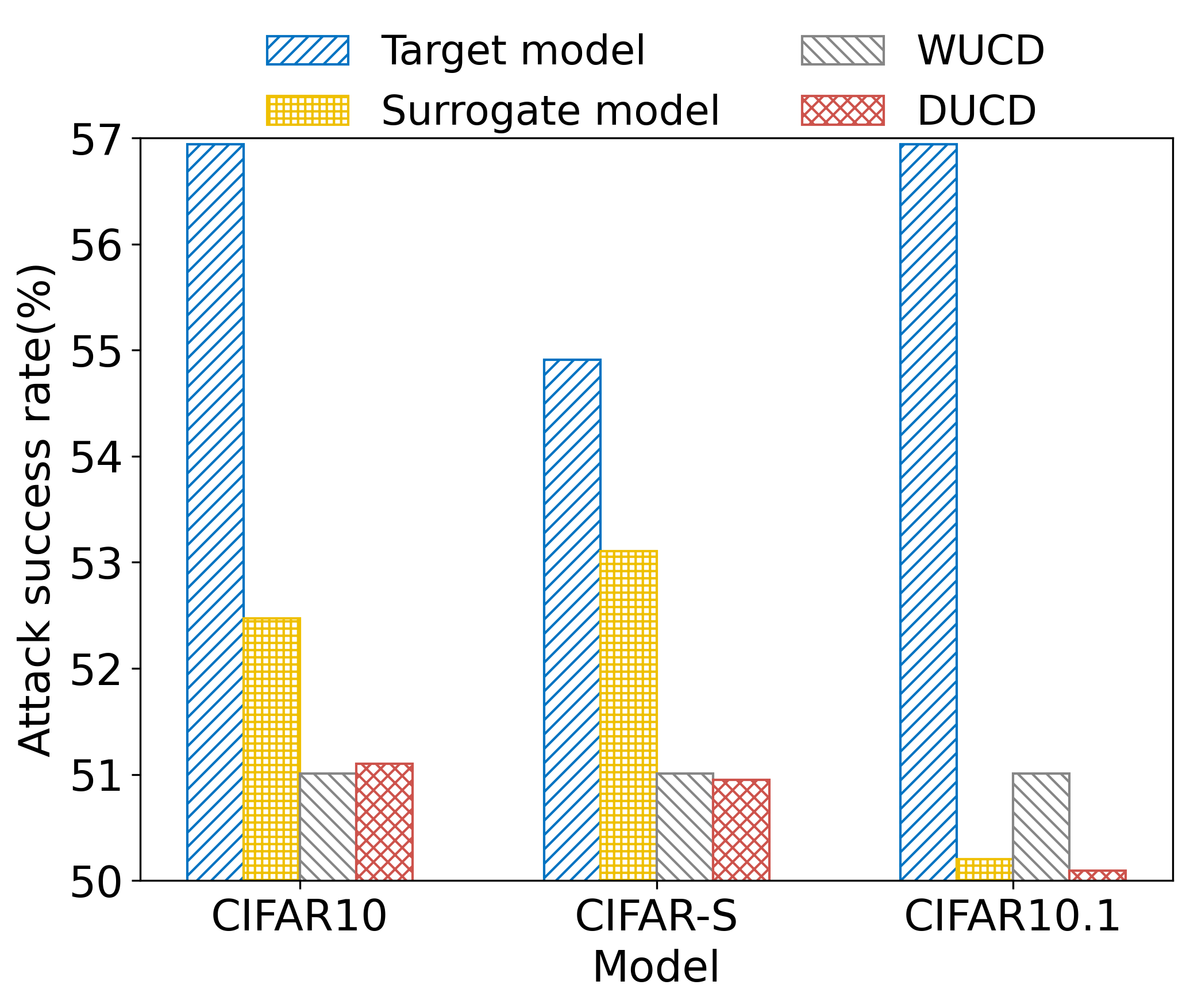} }\hspace{1mm}
    \subfloat[$\ell_\infty$]{\includegraphics[width=0.31\linewidth]{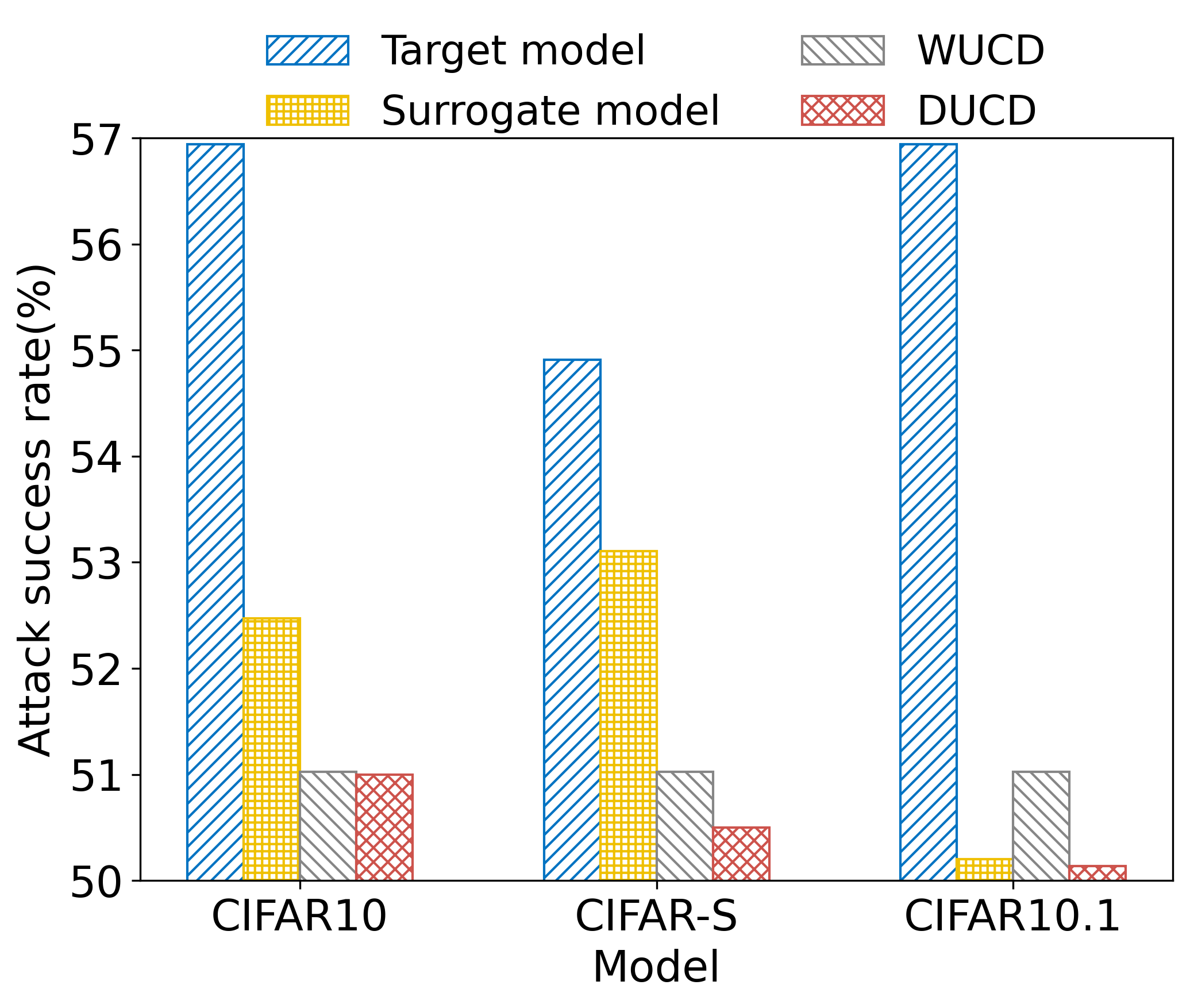} }
    \caption{Membership inference attack success rates against different models.}
    \label{fig: Membership-inference-attack-success-rate-under-different-models}
\end{figure*}



As shown in Fig. \ref{fig: Membership-inference-attack-success-rate-under-different-models}, the membership inference attack on the target model exhibits a high ASR of around 57\%, while the surrogate model derived from distillation significantly reduces it to around 52\%. Moreover, applying the certified defense to either the target model or the surrogate model also significantly reduces the ASR, which can be explained by the effectiveness of differential privacy ~\cite{senekane2017privacy}. Interestingly, when distillation and certification are conducted on a dataset different from the training dataset, the ASR is further reduced to less than 51\%, which is close to the success rate of random guessing.

\section{Related Work}
In this section, we brief existing works on empirical defense and certified defense.

\textbf{Empirical defense.}
Empirical defense is obtained by researchers through experimentation based on existing attacks. 
In practice, they exhibit good defensive performance against specific adversarial attacks.
Empirical defense includes adversarial detection and adversarial training.
Adversarial detection methods \cite{xufeature,geifman2019selectivenet, tian2021detecting} aim to identify and discard adversarial examples. While they ensure the robustness of model predictions, they don't enhance the model's intrinsic defense against attacks.
Adversarial training, on the other hand, improves DNNs' robustness by training them on datasets comprising both adversarial and normal examples. Pioneering research by Madry et al. \cite{madrytowards} theoretically framed this approach, leading to significant strides in empirical defense. Subsequent methods have varied in focus: some have optimized input \cite{songrobust}, others have refined training strategies \cite{dong2020adversarial, wang2020improving, jia2022adversarial, wang2023better, jin2023randomized}, and others still have explored the mechanisms underpinning adversarial training \cite{andriushchenko2020understanding}.
Although the methods mentioned above improve the robustness of the target model, they are designed based on existing attacks and therefore vulnerable to re-attacks specifically designed to bypass them.

\textbf{Certified defense.}
Certified defense is designed to achieve provable security for models, ensuring reliable robustness for machine learning classifiers against adversarial perturbations. A range of certified defense methods have been proposed.
Lecuyer et al. \cite{lecuyer2019certified} first introduced a certified defense that used differential privacy to offer robust proofs for classifiers. Cohen et al. \cite{cohen2019certified} later established a certified defense based on randomized smoothing, utilizing Gaussian noise to ensure rigorous robustness guarantees for the $\ell_2$ norm.
Subsequent works, including \cite{lee2019tight,teng2020ell_1,levine2020robustness, hammoudeh2024provable}, tailored their approaches to $\ell_p$-norm perturbations, but these methods are only robust for specific $\ell_p$-norm disturbances. Several recent studies \cite{chen2022input, nandi2023certified,wang2022improved,kumar2022towards} enhanced randomized smoothing to ensure model robustness within the $\ell_2$ and $\ell_\infty$ norms, albeit at the cost of a somewhat reduced robust radius.
While these methods successfully amplify model robustness and provide provable security, they are based on a white-box model assumption and rely heavily on prior knowledge of the model's architecture, parameters, and dataset. This dependence significantly hampers the practical application of the certified defense.

 
In addition to the above, researchers have developed several black-box certified defense methods. Salman et al.~\cite{salman2020denoised} pioneered an approach using a proxy model to approximate the black-box model, constructing the defense strategy around this surrogate. Carlini et al.~\cite{carlini2022certified} further enhanced this method with a denoising diffusion probability model, though its high training cost can be prohibitive for those with limited computational resources. However, these approaches~\cite{Nayak_2023_WACV} do not fully qualify as true black-box defenses, as they require prior knowledge of the target model's type and behavior. More recent studies~\cite{zhangrobustify, verma2023certified} propose certified defenses based solely on querying the target classifier. However, these methods are limited to defending against $\ell_2$ norm perturbations, lacking norm-universality.

\section{Conclusion}

This paper focuses on enhancing the robustness of pre-trained black-box models, aiming to reduce the strictness of defense assumptions and explore how to provide a universal defense mechanism that protects privacy and defends against all norm attacks through query operations alone. To achieve this goal, we propose an innovative method that does not rely on the internal details of the model. Instead, it uses query results to construct a universal defense system capable of resisting various adversarial attacks. Specifically, we design a knowledge distillation-based surrogate model generation technique to create a proxy model functionally similar to the target model. Combined with optimized randomized smoothing, this proxy model generates a robust certified classifier with a larger certified radius. This classifier effectively defends against adversarial attacks of any norm and maintains privacy protection.

Although our proposed method achieves the effectiveness of white-box certified defenses, it performs poorly against $\ell_\infty$ attacks. Even with rigorous theoretical derivations and proofs, defenses based on randomized smoothing is theoretically capable of defending against attacks of all norms, including $\ell_\infty$. However, our experiments demonstrate that in the face of real adversarial attacks, randomized smoothing fails to effectively defend against $\ell_\infty$ norm attacks. This remains a critical issue for future exploration and resolution in certified defenses.

\bibliographystyle{IEEEtran}
\bibliography{sample-base}

\vspace{-5mm}

\vfill

\end{document}